%% file: arxiv_bytedance.tex
\documentclass[]{bytedance}
\usepackage[toc,page,header]{appendix}

\usepackage{amsmath}
\usepackage{amsfonts}
\usepackage{amssymb}
\usepackage{mathtools}
\usepackage{xspace}

\usepackage{graphicx}
\usepackage{subcaption}
\usepackage{caption}
\usepackage{float}
\usepackage{wrapfig}

\usepackage{multirow}
\usepackage{colortbl}
\usepackage{arydshln}

\usepackage{algorithm}
\usepackage{algorithmicx}
\usepackage{algpseudocode}

\usepackage{hyperref}
\usepackage{url}

\usepackage{booktabs}
\usepackage{xcolor}
\newcommand{\improve}[1]{\textcolor{gray}{\small~(+#1)}}

\input{math_commands.tex}

\newcommand{\methodname}{PairUni}

\renewcommand{\cite}[1]{\citep{#1}}

\title{\protect\methodname: Pairwise Training for Unified Multimodal Language Models}

\author{
Jiani Zheng, 
Zhiyang Teng, 
Kunpeng Qiu, \\
Xiangtai Li, 
Anran Wang, 
Yu Tian, 
Ye Tian, 
Haochen Wang, 
Zhuochen Wang
}

\affiliation[]{ ByteDance }

\abstract{
Unified Vision-Language Models (UVLMs) perform both understanding and generation within a single architecture.
Since these models rely on heterogeneous data and supervision, balancing both generation and understanding in reinforcement learning (RL) is challenging.
To address this challenge, we propose \methodname, a unified framework that reorganizes data into understanding–generation (UG) pairs and aligns optimization accordingly. 
Specifically, we construct a unified paired dataset by synthesizing aligned instances via cross-modal semantic completion and retrieving semantically related samples. These paired structures expose cross-task semantic correspondences and support consistent policy learning. 
To leverage this structure, we present PairGRPO, a pair-aware variant based on Group Relative Policy Optimization. 
It assigns a similarity score to each pair to modulate the advantage, strengthening learning from well-aligned examples and reducing task interference.
Extensive experiments across diverse UVLM architectures (Autoregressive and Discrete Diffusion) and scales (1B to 14B) demonstrate that \methodname\ yields consistent improvements over strong baselines. 
Notably, our method also demonstrates strong generalization by improving performance on image editing tasks without using any editing-specific data.
Code: \url{https://github.com/Haochen-Wang409/PairUni}
}

\date{\today}
\correspondence{ 
\email{zhengjiani.0123@bytedance.com},
\email{zhiyang.teng@bytedance.com},
}

\begin{document}

\maketitle

\input{sections/1-intro}
\input{sections/3-method}
\input{sections/4-exp}
\input{sections/5-conclusion}

\bibliography{main}
\bibliographystyle{plainnat}

\newpage
\beginappendix

\input{sections/appendix_overview}
\input{sections/appendix_related_work}

\input{sections/appendix_data_details}

\input{sections/appendix_ablation}
\input{sections/appendix_more_models}
\input{sections/appendix_case_studies_understanding}
\input{sections/appendix_prompts}

\end{document}

%% file: math_commands.tex
\usepackage{amsmath,amsfonts,bm}

\def\eqref#1{equation~\ref{#1}}

\def\1{\bm{1}}

\DeclareMathAlphabet{\mathsfit}{\encodingdefault}{\sfdefault}{m}{sl}
\SetMathAlphabet{\mathsfit}{bold}{\encodingdefault}{\sfdefault}{bx}{n}



%% file: sections/1-intro.tex
\section{Introduction}
\label{sec:intro}

UVLMs have demonstrated strong performance in both multimodal understanding and image generation tasks with different architectures~\cite{lu2024ovis,chen2025blip3,deng2025bagel,lumina-dimoo}. 
However, as evaluation protocols increasingly emphasize complex, multi-step reasoning—spanning mathematics, the natural sciences, and multi-hop visual question answering—the objective of training a single system that simultaneously balances these capabilities within a unified learning paradigm remains highly challenging~\cite{li2023evaluating,chen2024we,lu2023mathvista}.%
This challenge is particularly acute during reinforcement learning (RL) stages, since understanding and generation are supervised with heterogeneous objectives and data formats, making the optimization process highly sensitive to data batching and cross-task credit assignment.

This fundamental disparity has largely led to various advancements.
For instance, text-to-image RL has focused on improving object controllability and prompt adherence \cite{jiang2025t2i,pan2025unlockingahamomentsreinforcement}, while visual reasoning improvements have separately targeted accuracy on benchmarks for math or science.
Consequently, attempts at unified RL encounter significant practical obstacles: (i) task interference during joint optimization, where gains on one objective cause regressions on the other \cite{jiang2025coreinforcementlearningunifiedmultimodal}; (ii) a broad and diverse understanding task space (e.g., math, charts, OCR) that resists a single reward design; and (iii) limited guidance on how to select and align data for unified RL at scale, constraining both stability and ceiling performance \cite{hong2025reinforcing}.

Prevailing GRPO-based strategies often sidestep the central issue by focusing on a single capability, such as using understanding signals to improve generation quality \cite{jiang2025t2i}, or by adopting multi-stage schemes that alternate between tasks to find a fragile balance \cite{jiang2025coreinforcementlearningunifiedmultimodal}.
While effective to a degree, these methods do not directly tackle the core source of conflict: the lack of data-level semantic alignment between understanding and generation supervision and the absence of an optimization rule that respects this alignment. 
As a result, the shared policy is driven by competing gradients from unrelated signals, leading to unstable updates and uneven performance gains across tasks. This optimization-level conflict is not merely theoretical; it is empirically measurable and provides the direct motivation for our work.

\input{figs/data_pipeline}

We address this with \methodname, a simple yet unified RL framework that aligns the problem at both the data and optimization levels. 
On the data side, we reorganize heterogeneous supervision into understanding–generation (UG) pairs centered on the same or closely related images. Two complementary pair types are constructed. \textbf{Aligned pairs} are formed by completing single-task samples into unified quadruples, using GPT-o3 to add the missing caption or prompt for understanding data or to synthesize a question–answer pair for generation data, so that both objectives share the same instance. 
We use clustering method to select representative high-quality medoids from the unified quadruples. \textbf{Retrieved pairs} link a generation sample to a semantically related understanding sample via similarity search over image embeddings, which expands coverage when exact matches are scarce. 
This paired view exposes cross-task correspondences, namely what to attend to for understanding and how to express it for generation, on related content rather than on unrelated batches. 

On the optimization side, we develop PairGRPO, a pair-aware variant of GRPO that modulates the advantage by pair similarity. 
Aligned pairs receive full weight, and retrieved pairs are down-weighted by their pair-similarity scores. 
This mechanism strengthens updates from high-quality supervision and tempers weaker matches, which reduces cross-task interference while preserving GRPO stability through the clipped objective and KL regularization. 
In effect, the policy update is made to respect the semantic alignment already present in the data.

Our contributions are summarized as follows:
\vspace{-0.5em}
\begin{itemize}
    \item We propose \methodname, a unified RL framework that organizes data into UG pairs and introduces PairGRPO to modulate optimization via similarity, effectively mitigating task interference.
    \item We construct PairUG-16k, a high-quality paired dataset specifically designed to support unified RL fine-tuning.
    \item Experiments show that \methodname\ achieves balanced improvements on Janus-Pro, outperforming strong baselines, and generalizes effectively to discrete diffusion models like Lumina-DiMOO.
\end{itemize}
\vspace{-0.5em}
These findings support a simple conclusion: pairing the data and weighting the advantage by pair-similarity is a general and effective ingredient for unified multimodal training, which improves understanding and generation together rather than trading one for the other.

%% file: figs/data_pipeline.tex
\begin{figure*}[!t]
    \centering
    \includegraphics[width=1.0\textwidth]{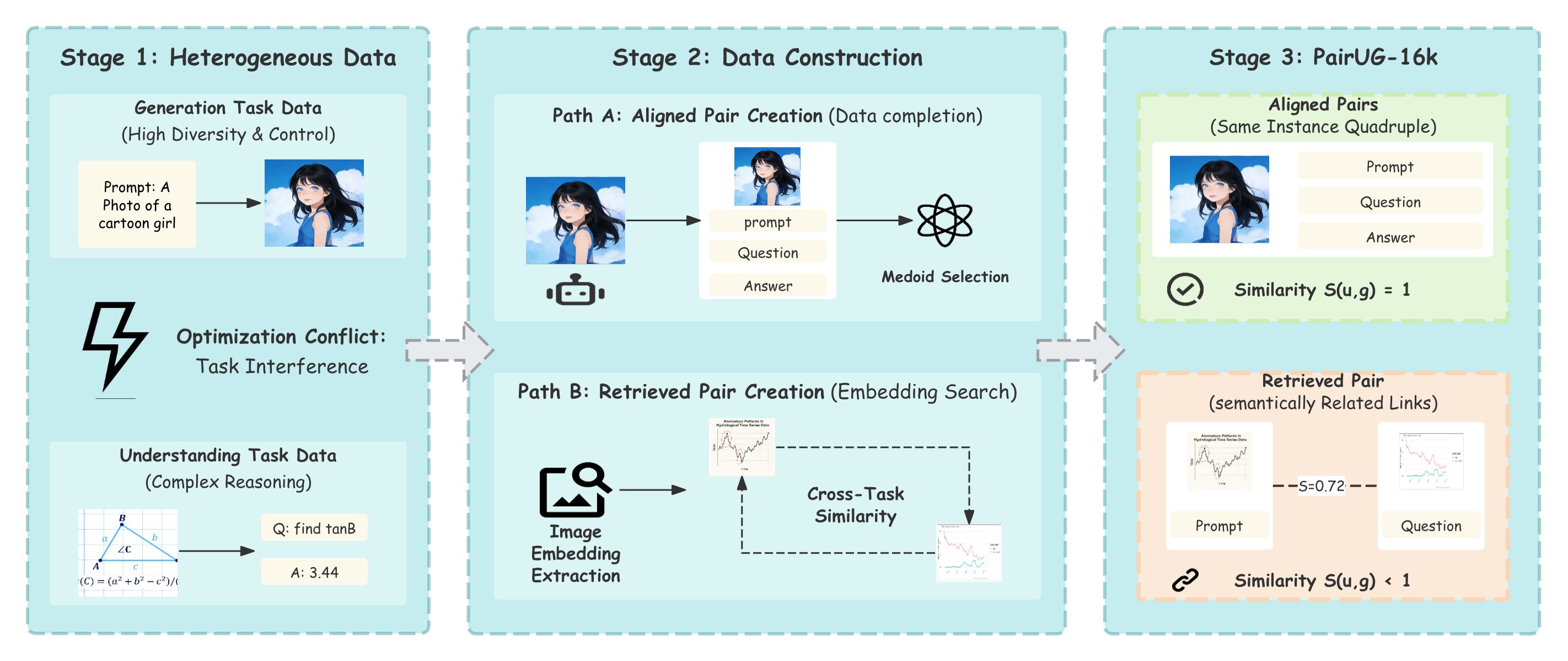}
    \caption{\textbf{Data Pairing Pipeline.} Left: Illustrative examples of aligned quadruples constructed from generation and understanding tasks. Right: Schematic of the proposed pairing strategy, which utilizes retrieval and clustering techniques to synthesize high-quality paired data (PairUG-16k) for unified training.}
    \label{fig:data_pipeline}
    \vspace{-1em}
\end{figure*}

%% file: sections/3-method.tex
\section{Method}

\label{sec:method}
\methodname\ has two key components: a data pairing pipeline (as shown in Figure~\ref{fig:data_pipeline}) to generate training pairs for unified models (Section~\ref{sec:data_pipeline}) and a PairGRPO algorithm (as shown in Figure~\ref{fig:framework} of Section~\ref{sec:method_pair_GRPO}), which is specially designed for RL of understanding-generation pairs.   

\subsection{Pairing Understanding–Generation Pairs}
\label{sec:data_pipeline}

We aim to construct a unified paired dataset $\mathcal{S} = \{(I, C, Q, A)\}$, where each data item supports both generation and understanding capabilities within a single multimodal model. $I$ is an image input,  $C$ is a text prompt that describes or motivates the image (used in generation), $Q$ is a visual understanding question and $A$ is the corresponding answer to $Q$. We build this paired dataset from two distinct sources: 1) \textbf{Understanding data} \( \mathcal{U} = \{(I, Q, A)\} \), where the image is annotated with comprehension questions; 2) \textbf{Generation data} \( \mathcal{G} = \{(I, C)\} \), where an image is paired with a generative prompt. These two sources are inherently heterogeneous and rarely aligned. To unify them, we design a data pipeline that constructs either \textit{aligned} or \textit{retrieval-based} pairs (see Appendix~\ref{app:data} for detailed statistics).

\subsubsection{Generating Aligned Pairs}
\input{tables/pipeline}
Given $U$ and $G$, we first augment each element in the original datasets to the desired quadruples and then we design an algorithm to select representative pairs to construct $\mathcal{S}$. 

\noindent\textbf{Cross-Modal Semantic Completion} We view the construction of aligned pairs not merely as augmentation, but as a \textit{semantic completion} process that bridges the information gap between tasks.
For understanding-only samples \( (I, Q, A) \in \mathcal{U} \), we employ GPT-o3 to generate the generation prompt \( C \) by conditioning on both the image $I$ and the QA pair $(Q, A)$. The instruction explicitly requires that $C$ must include the visual details necessary to answer $Q$ (e.g., if $Q$ asks about counting cats, $C$ must explicitly describe the number of cats).
Conversely, for generation-only samples \( (I, C) \in \mathcal{G} \), we use GPT-4o to generate $(Q, A)$ by conditioning on $C$, ensuring that the question targets the specific visual elements described in the generation prompt.
Crucially, we introduce a \textbf{consistency check} loop: after generation, we verify whether the synthesized component contradicts the original data (e.g., ensuring the generated $C$ does not hallucinate objects absent in $I$). This generate-then-verify pipeline ensures that the resulting quadruples $(I, C, Q, A)$ are strictly semantically bound, eliminating semantic drift and ensuring that reward signals for both tasks are grounded in the exact same visual context.

\noindent\textbf{Data Selection}  Even after augmentation, many samples may be redundant or low-quality.  To ensure coverage and diversity, we adopt a clustering-based strategy to identify {representative aligned pairs} ($\mathcal{D}_\text{aligned}$) from these datasets as shown in the first part of Algorithm~\ref{alg:data_pipeline_compact}. First, we extract image features using pretrained visual encoders, and then perform K-means clustering over the joint visual feature space of \(\mathcal{U} \cup \mathcal{G}\). Second, for each cluster, we select the most central sample (i.e., the medoid) as a canonical representation of that cluster's content. This yields a curated set of self-referential pairs where understanding and generation annotations coexist for the same image. These samples are both semantically rich and geometrically representative of the data distribution, forming a strong backbone for joint training.

\subsubsection{Constructing Retrieval-Based Pairs}
While aligned pairs are semantically precise, their quantity is limited. To scale supervision, we introduce \textbf{retrieval-based pairs} $\mathcal{D}_{\text{ret}}$, where understanding and generation samples come from \emph{different images} but share visual similarity. 
The second part of Algorithm~\ref{alg:data_pipeline_compact} shows the algorithm.  The main idea is to extract visually similar image pairs from two datasets to establish correspondences between ``understand'' and ``generate'' data.  First, cosine similarity is computed across all remaining generation–understanding image pairs. For each generation image, we retrieve the top-$n$ most similar understanding images above a similarity threshold $\delta$. A greedy matching algorithm is used to ensure that each image is only used once. This retrieval mechanism exploits the fact that \emph{semantically similar images often support related tasks}, even if not identical. By leveraging these approximate matches, the model can learn \emph{cross-instance generalization}, which enhances its robustness and expands training coverage.

Together, $\mathcal{D}_{\text{aligned}}$ and $\mathcal{D}_{\text{ret}}$ form the UG pair set $\mathcal{S}$ (referred to as \textbf{PairUG-16k}) used for policy optimization. These two pathways provide {complementary benefits}: aligned pairs deliver precise, high-quality supervision, while retrieval pairs enhance scale and semantic diversity, covering a wide spectrum of multimodal understanding and generation tasks.

\subsection{PairGRPO}
\label{sec:method_pair_GRPO}

This section describes: (1) vanilla GRPO with mixed tasks; (2) pairwise GRPO with UG‐pairs; (3) PairGRPO, which incorporates pair similarity into advantage weighting. The formulation is model-agnostic and can be applied to both autoregressive and discrete diffusion backbones. Each step is designed to better align understanding and generation, reduce conflict, and stabilize learning.

\input{figs/framework}
\noindent\textbf{(1) Vanilla GRPO with mixed tasks.}  
We consider a batch of trajectories $\mathcal{B} = \{\tau_i\}_{i=1}^N$ mixed with understanding and generation tasks. For each trajectory $\tau_i$ with input $q$ and output $o_{1:T}$, we compute the token-wise importance ratio $\rho_t(\theta) = \frac{\pi_\theta(o_t \mid q, o_{<t})}{\pi_{\theta_{\mathrm{old}}}(o_t \mid q, o_{<t})}$.
Rewards $r$ ($r_{\mathrm{Und}}$ or $r_{\mathrm{Gen}}$) are normalized within groups sharing the same prompt to obtain the advantage $\widehat A_t = (r - \mu_r) / \sigma_r$.
The objective maximizes the clipped surrogate loss:
{\small
\begin{equation}
J_{\mathrm{vanilla}}(\theta) = 
\mathbb{E}_{\tau\sim \pi_{\theta_{\mathrm{old}}}}
\left[
\sum_{t=1}^T \mathcal{L}_{\mathrm{clip}}(\rho_t, \widehat A_t)
\right] - \beta D_{\mathrm{KL}}(\pi_\theta || \pi_{\mathrm{old}}),
\end{equation}
}
where $\varepsilon$ is the clipping threshold and $\beta$ controls the KL penalty (default $\beta=0$).

\noindent\textbf{Reward Functions}
Our reward functions are tailored to the specific goals of understanding and generation. For understanding tasks, typically formulated as multiple-choice question answering, we use prediction accuracy as our reward, a standard metric that directly measures correctness: $r_{\text{Und}} = \text{Acc}(y_{\text{pred}}, y_{\text{true}})$.

For generation tasks, we employ the HPSv2 reward model~\cite{wu2023human} to evaluate output quality. The reward is given by:$r_{\text{Gen}} = R_{\text{HPSv2}}(x, y_{\text{gen}})$, where $x$ is the input prompt and $y_{\text{gen}}$ is the corresponding generated image.

\noindent\textbf{(2) Pairwise GRPO with UG data pairs.} 
To mitigate the task interference inherent in mixed-batch training (as discussed in Section~\ref{sec:intro}), we reorganize training around a set of paired training examples $\mathcal{P} = \{p\}_{p=1}^M$. Each pair $p$ consists of two datapoints: one generation example and one understanding example that are semantically aligned.  This pairing is defined at the data level, not at the trajectory level: each data item in the pair can produce multiple trajectories through sampling.

For each paired datapoint $p$, we generate a set of trajectories $\{\tau^{(u)}_{p,k}\}_{k=1}^{K_u}$ for the understanding side, and $\{\tau^{(g)}_{p,k}\}_{k=1}^{K_g}$ for the generation side. These trajectories are grouped by task type and prompt to compute group-relative advantages. Specifically, we calculate $\widehat A^{(u)}_t$ and $\widehat A^{(g)}_t$ as the normalized reward within the respective task-type group, using GRPO's group-based normalization.

The pairwise GRPO objective is then defined as:
\begin{equation}
\begin{aligned}
J_{\mathrm{pair}}(\theta) = \mathbb{E}_{p \sim \mathcal{P}} \Bigg[ \sum_{\tau \in \{\tau^{(u)}_p\}} \sum_{t \in \tau} \mathcal{L}_{\mathrm{clip}}\big(\rho_t, \widehat{A}_t^{(u)}\big) + \\
\sum_{\tau \in \{\tau^{(g)}_p\}} \sum_{t \in \tau} \mathcal{L}_{\mathrm{clip}}\big(\rho_t, \widehat{A}_t^{(g)}\big) \Bigg],
\end{aligned}
\label{eq:grpo-pairwise}
\end{equation}

This formulation ensures that policy gradients from understanding and generation are derived from semantically related training examples, even when multiple trajectories are sampled per side. This pairing encourages more consistent policy updates across tasks.

\noindent\textbf{(3) PairGRPO: similarity-weighted advantage.}
This design is motivated by an empirical observation: when understanding and generation data are semantically aligned, the cosine similarity between their gradients increases (see Figure~\ref{fig:intro} in Appendix~\ref{app:gradient_analysis}). Higher gradient agreement correlates with better unified performance, suggesting that weighting updates by semantic alignment can reduce task interference. To further align training strength with semantic similarity, we introduce a pair-level similarity score $s_p \in [0, 1]$ for each data pair $p$, computed via a pretrained image encoder. Based on this, we assign a pair weight $w_p$:
\begin{equation}
\small
w_p =
\begin{cases}
1, & \text{if } p \in \mathcal{D}_{\text{aligned}}, \\
\sqrt{s_p}, & \text{if } p \in \mathcal{D}_{\text{ret}}
\end{cases}
\end{equation}
and apply this weight to all advantages computed from trajectories originating from the pair:
\[
\widetilde A^{(u)}_t = w_p\,\widehat A^{(u)}_t,
\qquad
\widetilde A^{(g)}_t = w_p\,\widehat A^{(g)}_t.
\]

We use the square root to amplify the relative differences between similarity scores, as all selected pairs are drawn from a high-similarity candidate pool. The full PairGRPO objective becomes:
\begin{equation}
\begin{aligned}
J_{\mathrm{PairUni}}(\theta) = \mathbb{E}_{p \sim \mathcal{P}} \Bigg[ \sum_{\tau \in \{\tau^{(u)}_p\}} \sum_{t \in \tau} \mathcal{L}_{\mathrm{clip}}\big(\rho_t, \widetilde{A}_t^{(u)}\big) + \\ \sum_{\tau \in \{\tau^{(g)}_p\}} \sum_{t \in \tau} \mathcal{L}_{\mathrm{clip}}\big(\rho_t, \widetilde{A}_t^{(g)}\big) \Bigg].
\end{aligned}
\label{eq:pairgpro}
\end{equation}

This design modulates the trajectory-level credit assignment based on the quality of semantic pairing, strengthening updates from well-aligned pairs (aligned: $w_p=1$) while attenuating noisy or weakly aligned ones (retrieved: $w_p=\sqrt{s_p}$). By co-optimizing with the structure of PairUG-16k, PairGRPO retains the stability of GRPO while effectively leveraging data-level semantic alignment to resolve optimization-level task conflicts.

%% file: tables/pipeline.tex
\begin{wrapfigure}{r}{0.5\linewidth}
\vspace{-0.8em}
\captionsetup{type=algorithm}
\caption{Pseudocode for the Data Pairing Algorithm.}
\label{alg:data_pipeline_compact}
\small
\begin{algorithmic}[1]
\State \textbf{Input:} features $\mathcal{F}_u,\mathcal{F}_g$, clusters $K$, neighbors $n$
\State $\mathcal{F}\!\gets\!\mathrm{L2Norm}(\mathcal{F}_u\!\cup\!\mathcal{F}_g)$
\State $\mathcal{C}\!\gets\!\mathrm{MiniBatchKMeans}(\mathcal{F},K)$
\For{$k=1..K$}
  \State $\mathcal{I}_k\!\gets\!\{i:\mathrm{assign}(x_i)=k\}$
  \State $i^*\!\gets\!\arg\max_{i\in\mathcal{I}_k}\langle f_i,c_k\rangle$
  \State $\mathcal{D}_{\text{aligned}}\!\gets\!\mathcal{D}_{\text{aligned}}\cup\{(x_{i^*},y_{i^*})\}$
\EndFor
\State $\mathcal{F}_u^{\rm rem}\!\gets\!\mathcal{F}_u\!\setminus\!\mathcal{D}_{\text{aligned}}$;\ 
       $\mathcal{F}_g^{\rm rem}\!\gets\!\mathcal{F}_g\!\setminus\!\mathcal{D}_{\text{aligned}}$
\For{$x_i^g\in\mathcal{F}_g^{\rm rem}$}
  \State $\mathcal{J}\!\gets\!\operatorname{top}\!-\!n\ \mathrm{kNN}(\phi_i^g,\mathcal{F}_u^{\rm rem})$
  \State $\mathcal{D}_{\text{ret}}\!\gets\!\mathcal{D}_{\text{ret}}\cup\{(x_{j}^u,x_i^g): j\!\in\!\mathcal{J}\}$;\ 
        \State $\mathcal{F}_u^{\text{rem}} \gets \mathcal{F}_u^{\text{rem}} \setminus \mathcal{J}$, $\mathcal{F}_g^{\text{rem}} \gets \mathcal{F}_g^{\text{rem}} \setminus \{x_i^g\}$
\EndFor
\State \textbf{Output:} $\mathcal{D}_{\text{aligned}}\cup\mathcal{D}_{\text{ret}}$
\end{algorithmic}
\end{wrapfigure}

%% file: figs/framework.tex
\begin{figure*}[t]
    \centering
    \includegraphics[width=1.0\textwidth]{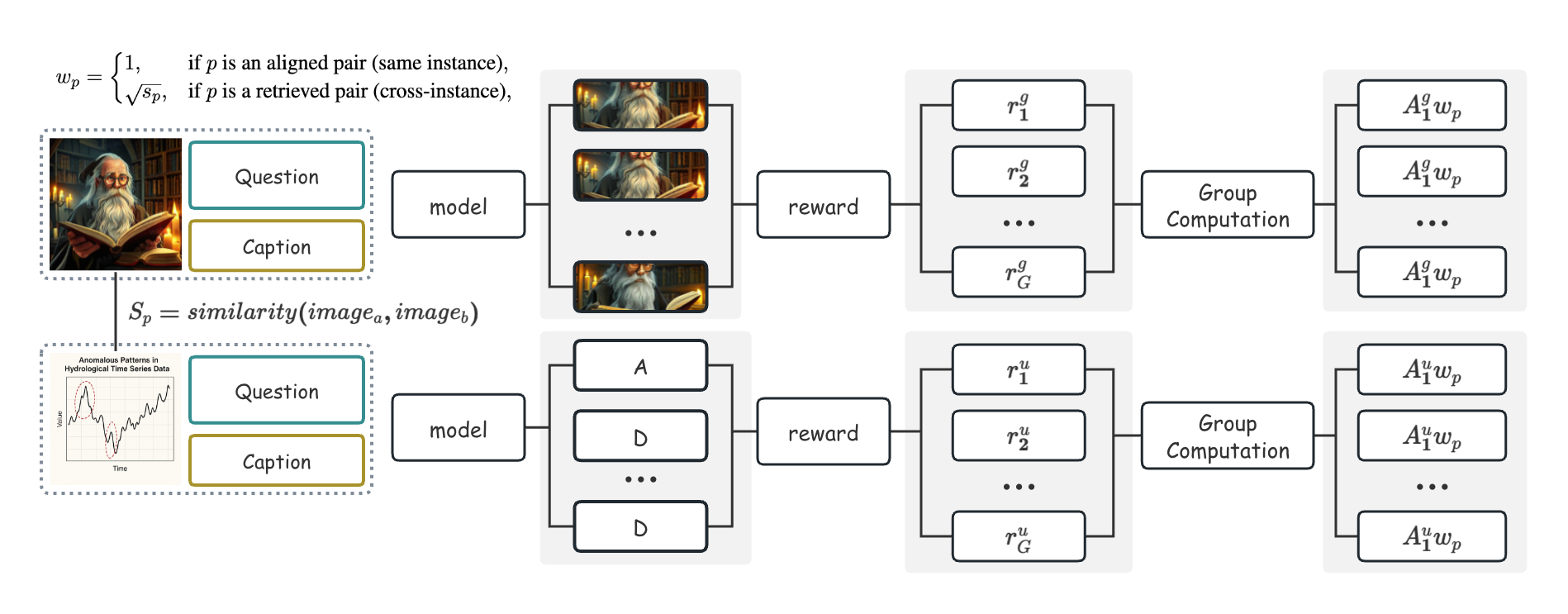}
    \caption{\textbf{Framework of PairUni}: Overview of the proposed architecture, featuring a dual-component design that integrates a sophisticated data processing pipeline with the GRPO reinforcement learning algorithm to achieve unified multimodal understanding and generation.}
    \label{fig:framework}
    \vspace{-1em}
\end{figure*}

%% file: sections/4-exp.tex
\section{Experiments}

\noindent\textbf{Training} 
We adopt Janus-Pro~\cite{chen2025janusprounifiedmultimodalunderstanding} as a primary baseline because it is widely used as a comparator for unified multimodal understanding and generation and exhibits competitive performance. 
All experiments are conducted on \(8\times\) H100 GPUs. For the 7B model, we use a rollout size of 4 for both text and image generation and train for at most 1200 steps. The per-device batch size is 2 (global batch size \(16\)). We set the classifier-free guidance (CFG) weight to \(5\), \(\beta=0\), the learning rate to \(1\times 10^{-6}\), and the sampling temperature to \(1.0\). 
For the 1B model, the rollout size for both modalities is increased to \(8\). All the visual features are extracted using a ResNet50 encoder by removing the classification head~\cite{koonce2021resnet} and L2-normalized (we compare different image encoders in Appendix~\ref{app:ablation}). 
We use the Orsta data~\cite{ma2025one} as the understanding data $\mathcal{U}$, which contains about 47K samples, and the BLIP3o data~\cite{chen2025blip3} as the generation dataset $\mathcal{G}$, which contains about 60K samples. 
We exclude the original detection and grounding QA pairs from Orsta since Janus-Pro fails on these tasks. 
The similarity threshold is 0.6. Our constructed PairUG-16k dataset consists of 16,320 samples: including 4,971 aligned pairs and 11,349 retrieval-based pairs. The similarity score of each pair is provided (see Appendix~\ref{app:data} for dataset details). 

\noindent\textbf{Evaluation}  
We evaluate multimodal understanding on MME Perception (basic reasoning), POPE~\cite{li2023evaluating} (hallucination), MMStar~\cite{chen2024we} (visual reasoning), and MMMU~\cite{yue2024mmmu} (multi-discipline tasks).
For generation, we use GenEval~\cite{ghosh2023geneval} to measure visual fidelity and WISE~\cite{niu2025wise} to assess knowledge-grounded synthesis.

\subsection{Main Results}
\label{sec:main_results}

We first present the main results on Janus-Pro 1B and 7B, demonstrating that \methodname\ effectively addresses the trade-off typically observed in unified modeling, leading to simultaneous improvements in both understanding and generation.

\input{tables/understand}
\input{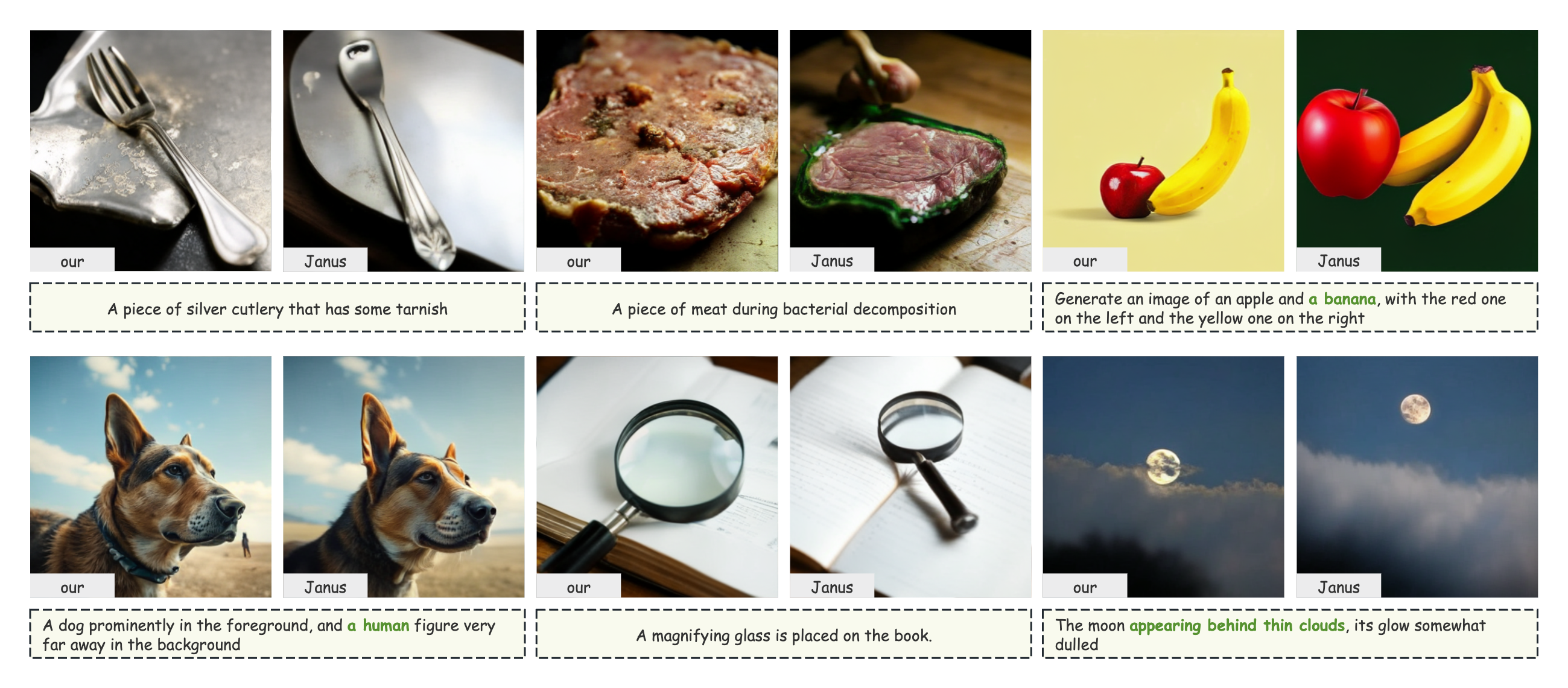}

\noindent\textbf{Multimodal Understanding.} 
Table~\ref{tab:multimodal_selected} reports the performance of different models on four representative multimodal understanding benchmarks. 
On the broad-coverage {MMMU} benchmark, which emphasizes general visual reasoning across scientific, mathematical, and commonsense domains, \methodname\ achieves substantial improvements among unified models: at the 1B scale it attains a score of {40.4} (cf.\ Janus-Pro-1B: 36.3; ULM-R1$^{\dagger}$: 40.3), and at the 7B scale it reaches {47.0}, surpassing prior unified baselines such as Janus-Pro-7B (41.1) and UniToken (32.8). 
Beyond MMMU, \methodname\ also improves performance on {MMStar} and {MME}: at 7B, it increases MMStar from 46.5 to {49.5} and MME from 1567.1 to {1597.7}, while maintaining competitive {POPE} performance (88.0 vs.\ 87.4), thereby demonstrating gains on both perception-intensive and knowledge-intensive tasks.
At the 1B scale, \methodname\ attains \textbf{46.4} on MMStar and \textbf{1483.2} on MME with stable POPE performance (86.4), indicating that the proposed data pairing strategy and PairGRPO optimization remain effective under stringent capacity constraints. 
These consistent improvements suggest that \methodname\ enhances unified visual reasoning capabilities without compromising perceptual robustness.

\input{tables/wise}
\input{tables/geneval}
\noindent\textbf{Text-to-Image Generation.} 
We evaluate text-to-image generation on the WISE~\cite{niu2025wise} and GenEval~\cite{ghosh2023geneval} benchmarks.  
On WISE, as shown in Table~\ref{tab:wise}, \methodname\ achieves the highest overall performance among unified models at both parameter scales, with scores of {0.38} (1B) and {0.45} (7B). At the 1B scale, it improves over Janus-1B (0.23) and ULM-R1 (0.33), and at the 7B scale it surpasses Janus-Pro-7B (0.35) and Emu3 (0.39), thereby substantially reducing the performance gap with generate-only models. Because ULM-R1~\cite{jiang2025coreinforcementlearningunifiedmultimodal} is not open-sourced, we report its performance only after Unified-RL and Refined-RL training on approximately 40k samples. At the subtask level, \methodname\ notably enhances performance on \emph{Space} (0.56 at 1B; 0.62 at 7B) and \emph{Physics} (0.44 at 1B; 0.55 at 7B), indicating improved grounding in geographic and physical commonsense.  

On \emph{GenEval} (Table~\ref{tab:gen_eval_results}), \methodname\ demonstrates strong compositional generalization at both scales. At 1B, \methodname-1B attains the highest overall score (0.79), exceeding Janus-Pro-1B (0.73) and ULM-R1 (0.76). At 7B, \methodname-7B achieves 0.85, outperforming Janus-Pro-7B (0.79) and DSR (0.84), the latter exhibiting comparatively weaker semantic understanding and a substantially underperforming 1B variant. 
Although Janus-Pro-R1 slightly outperforms \methodname-7B by a margin of one point (0.86), its understanding metrics and WISE scores are not reported.  

These improvements are consistent with our gains on WISE and suggest that PairGRPO more effectively enforces constraint adherence and object–attribute binding. 
Collectively, the results indicate that \methodname\ yields gains on \emph{both} sides of the unified objective, in contrast to baseline models that typically specialize in either understanding or generation, but not both (e.g., InternVL3 for understanding, FLUX.1 for generation). 
Additional qualitative examples are provided in Appendix~\ref{app:understanding-cases}.
These results not only validate the efficacy of our proposed framework but also highlight the potential of unified learning paradigms to surpass specialized models in complex multimodal tasks.

\noindent\textbf{Generalization Across Architectures and Scales.}
To demonstrate that \methodname\ captures a general principle of unified learning rather than being specific to one architecture, we evaluate its effectiveness across different model families and scales.
First, we test \methodname\ on \emph{Lumina-DiMOO}, a discrete diffusion model that differs fundamentally from the autoregressive Janus-Pro. As shown in Table~\ref{tab:more_model_performance} (a), \methodname\ yields consistent gains: {MMMU} improves from 58.6 to {61.3} (+2.7), {MMStar} from 52.4 to {52.6} (+0.2), and {GenEval} from 0.88 to {0.89}.
Second, we verify scalability on \emph{Bagel}, a larger 14B unified model. As presented in Table~\ref{tab:more_model_performance} (b), \methodname\ outperforms the random pairing baseline on Bagel-14B, boosting MMMU to 54.0 and POPE to 89.21, while also improving generation quality. 
These results, combined with our main experiments on Janus-Pro 1B and 7B, confirm that \methodname\ is effective across diverse architectures and scales ranging from 1B to 14B. More training details for these models are provided in Appendix~\ref{app:more_model}.

\noindent\textbf{Zero-shot Improvement in Editing Capabilities.}
A key hypothesis is that bidirectional alignment should enhance the model's ability to follow complex instructions, such as image editing, even in the absence of specific editing training data. We evaluate this zero-shot capability on Lumina-DiMOO and Bagel using image editing benchmarks. As shown in Table~\ref{tab:more_model_performance}, \methodname\ consistently improves performance across all editing subtasks (Add, Replace, Remove, Style). For instance, on Lumina-DiMOO, \methodname\ achieves higher scores in every category compared to the random pairing baseline. This demonstrates that our method inherently improves the model's capacity for instruction-following tasks by establishing better alignment between understanding and generation representations.

\subsection{Analysis: Effectiveness of PairUni Framework}
\label{sec:analysis}

To understand the source of our improvements, we conduct a series of ablation studies to isolate the impact of two core components: the data pairing strategy and the similarity-based advantage adjustment. We further analyze gradient similarity in Appendix~\ref{app:gradient_analysis}. We intervene on the semantic alignment between understanding and generation training signals by constructing six data-combination scenarios (PairUG, retrieval-based pairs, low-similarity unpaired data, understanding-only, generation-only, and random pairs), while keeping the training recipe fixed. As shown in Fig. 8(a), stronger semantic alignment consistently increases the median cosine similarity between the gradients from the understanding and generation objectives (e.g., 0.120 for PairUG vs. 0.059 for random/generation-only), indicating reduced gradient interference. Importantly, Fig. 8(b) shows that higher gradient agreement correlates with better joint downstream outcomes: PairUG achieves the best simultaneous performance on MMMU/MMStar and GenEval (40.4/46.1/79), whereas low-similarity or random pairing yields lower gradient agreement and inferior joint results (e.g., GenEval drops to 71 under low-similarity unpair). This controlled trend directly supports our motivation that aligned U–G pairs mitigate gradient conflicts and improve unified generalization.

\noindent\textbf{Disentangling Data Pairing from Data Quality.} 
A central question is whether the observed performance improvements arise solely from the use of higher–quality data or from the \emph{pairing} mechanism itself. 
Table~\ref{tab:data} addresses this by comparing our method with random pairing strategies under an identical data source and budget. 
\input{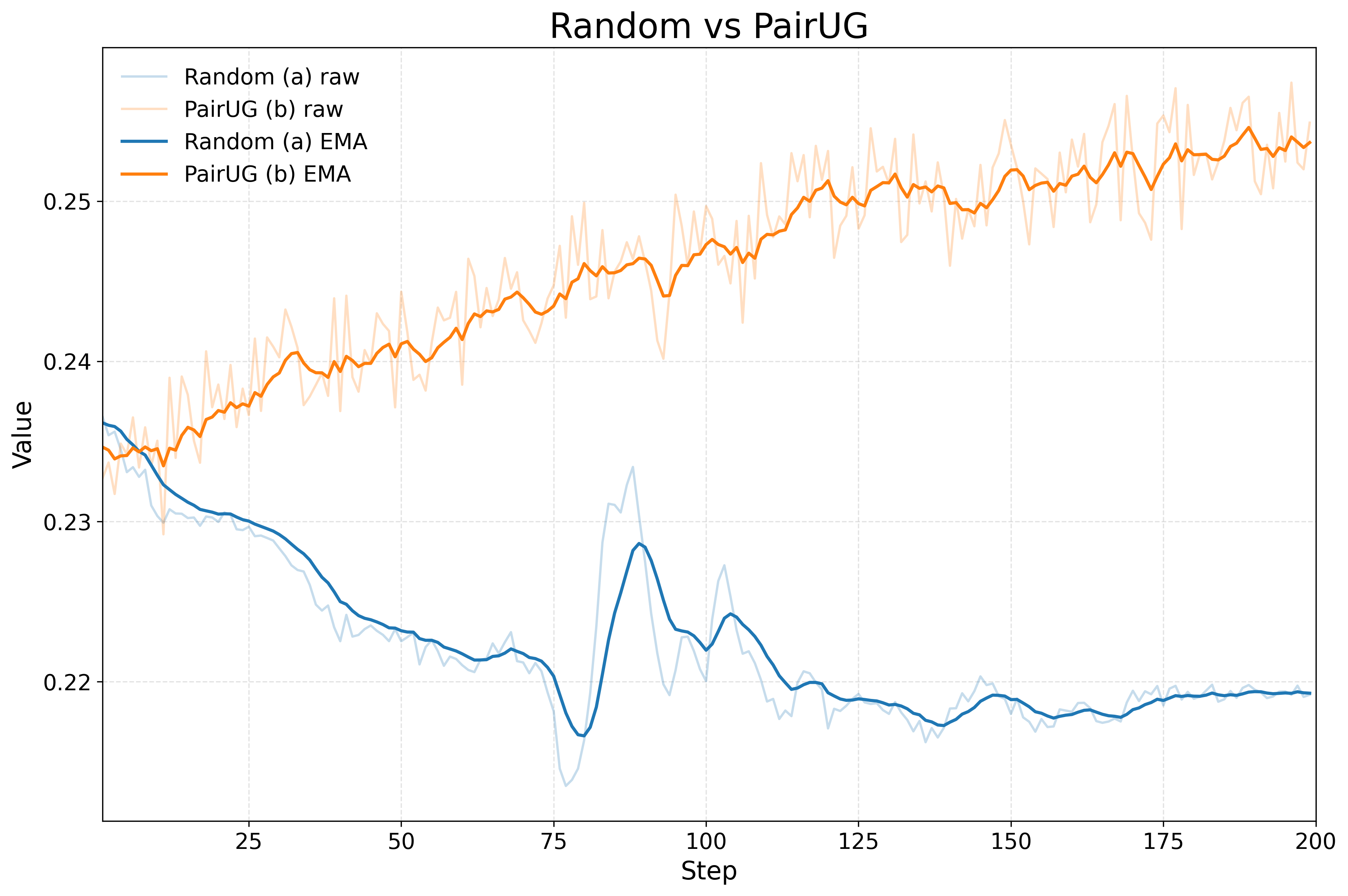}
Naive mixtures (\emph{unpaired} or \emph{random} pairing) under the same computational and data constraints systematically reduce generative fidelity (e.g., 0.71–0.73 on GenEval). 
This phenomenon is theoretically supported by our gradient analysis in Appendix~\ref{app:gradient_analysis} (Figure~\ref{fig:intro}): gradients induced by semantically unrelated (randomly paired) samples exhibit low cosine similarity, effectively behaving as conflicting noise that destabilizes the shared policy. In contrast, our aligned and retrieved pairs maintain higher gradient agreement, fostering constructive interference.
Furthermore, the construction of our dataset relies on selecting representative samples. As shown in Appendix~\ref{app:ablation} (Table~\ref{tab:ablation_kmeans}), our K-means based medoid selection strategy significantly outperforms random selection, ensuring that the curated dataset \textsc{PairUG-16k} covers diverse and high-quality semantic regions, which is crucial for the consistent improvements observed.
Figure~\ref{fig:reward_curve} further demonstrates that PairUG-16k yields more stable training dynamics relative to random pairing.

\input{tables/more_model_performance}
\vspace{0.5\baselineskip}
\input{tables/ablation_pair}

\noindent\textbf{Effect of Similarity-based Advantage Adjustment.} 
Table~\ref{tab:sim} isolates the specific contribution of similarity-based weighting within PairGRPO. 
We perform a detailed investigation of weighting strategies in Appendix~\ref{app:ablation} (Table~\ref{tab:ablation_weighting}). Our results show that the square-root scaling ($\sqrt{s_p}$) yields the strongest overall performance compared to linear scaling or no weighting. This non-linear scaling effectively balances the contribution of retrieved pairs: it down-weights weakly aligned pairs to prevent noise injection while preserving sufficient gradient magnitude for effective learning.
Additionally, the choice of similarity threshold is critical for the quality-quantity trade-off. As detailed in Appendix~\ref{app:ablation} (Table~\ref{tab:ablation_threshold}), a threshold of 0.6 is optimal; it ensures sufficient data volume for training while excluding low-similarity pairs that could degrade performance.
We also justify our choice of image encoder for similarity computation. In Appendix~\ref{app:image_encoder} (Table~\ref{tab:image_features}), we compare different extractors and find that those emphasizing visual similarity (e.g., ResNet, DINOv3) outperform high-level semantic encoders (e.g., Perception Encoder), confirming that visual consistency is the primary driver for effective retrieval-based pairing.
In summary, similarity-based weighting provides a principled mechanism to enhance comprehension-oriented performance while preserving generation quality by prioritizing supervision from strongly aligned pairs.

%% file: tables/understand.tex
\begin{table*}[t]
\centering
\small
\caption{\textbf{Main Results on multimodal understanding benchmarks.} Comprehensive comparison of \protect\methodname\ against state-of-the-art methods across various tasks and model sizes. The \protect\methodname\ method consistently achieves superior performance, validating the effectiveness of the unified approach.}
\scalebox{0.8}{
\begin{tabular}{l l l c c c c}
\toprule
\textbf{Model} & \textbf{LLM} & \textbf{MMMU} & \textbf{MMStar} & \textbf{MME(P)} & \textbf{POPE} \\
\midrule
\rowcolor{gray!15}
\multicolumn{7}{l}{Understanding Only} \\
InternVL3~\cite{zhu2025internvl3exploringadvancedtraining} & Qwen2.5-1.5B~\cite{Qwen2.5-VL} & 48.6 & \textbf{60.7} & - & \textbf{89.6} \\
Qwen2.5-VL~\cite{Qwen2.5-VL} & Qwen2.5-3B &\textbf{51.2} & 56.3 & - & 85.9 \\
LMM-R1~\cite{peng2025lmmr1empowering3blmms} & Qwen2.5-3B & - & 58.0 & - & - \\
\midrule
\rowcolor{gray!15}
\multicolumn{7}{l}{Unified Understanding and Generation} \\
Show-o\cite{xie2024showo} & Phi-1.3B~\cite{gunasekar2023textbooks} & 26.7 & - & - & 80.0 \\
HermesFlow~\cite{yang2025hermesflowseamlesslyclosinggap} & Phi-1.3B & 28.3 & - & - & 81.4 \\
Janus-Pro-1B~\cite{chen2025janusprounifiedmultimodalunderstanding} & DeepSeek-LLM-1.5B~\cite{deepseekai2025deepseekr1incentivizingreasoningcapability} & 36.3 & - & 1444.0 & 86.2 \\
$ULM-R1^{\dagger}$~\cite{jiang2025coreinforcementlearningunifiedmultimodal} & DeepSeek-LLM-1.5B &  40.3 & - & - & - \\
\rowcolor{cyan!5}
\textbf{\protect\methodname-1B} & DeepSeek-LLM-1.5B & \textbf{40.4} & \textbf{46.4} & \textbf{1483.2} & \textbf{86.4} \\
\hdashline
Orthus~\cite{kou2025orthusautoregressiveinterleavedimagetext} & Chameleon-7B~\cite{team2024chameleon} & 28.2 & - & 1265.8 & 79.6 \\
VILA-U~\cite{wu2025vilauunifiedfoundationmodel} & LLaMA-2-7B~\cite{touvron2023llama2openfoundation} & - & - & - & 85.8 \\
UniToken~\cite{jiao2025unitokenharmonizingmultimodalunderstanding} & Chameleon-7B & 32.8 & 46.1 & - & - \\
Janus-Pro-7B~\cite{chen2025janusprounifiedmultimodalunderstanding} & DeepSeek-LLM-7B & 41.1 & 46.5 & 1567.1 & 87.4 \\
DSR~\cite{hong2025reinforcing} & DeepSeek-LLM-7B & 41.1 & - & - & 86.6 \\
\rowcolor{cyan!5}
\textbf{\protect\methodname-7B} & DeepSeek-LLM-7B & \textbf{47.0} & \textbf{49.5} & \textbf{1597.7} & \textbf{88.0} \\
\bottomrule
\end{tabular}}
\label{tab:multimodal_selected}
\vspace{-1em}
\end{table*}

%% file: figs/cases.tex
\begin{figure*}[!t]
    \centering
    \includegraphics[width=0.9\textwidth]{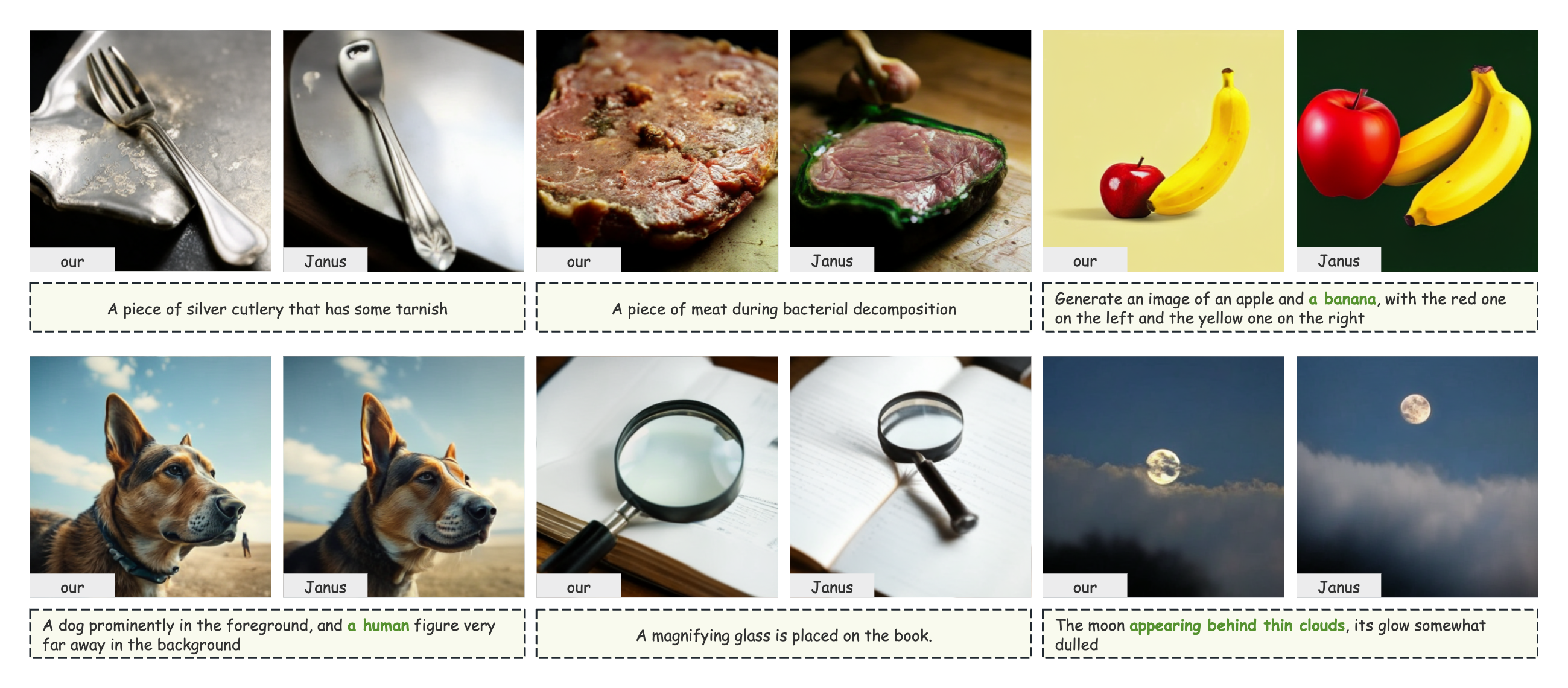}
    \caption{\textbf{Case Study}: Qualitative comparison of generated images between Janus-Pro-7B and \protect\methodname. The results demonstrate \protect\methodname's superior ability to follow complex prompts and generate high-fidelity images.}
    \label{fig:case}
    \vspace{-1em}
\end{figure*}

%% file: tables/wise.tex
\begin{table*}[t]
\centering
\caption{\textbf{Results on the WISE Benchmark.} \protect\methodname\ achieves the highest overall score among compared methods, with particularly outstanding performance in the space and physics subtasks, demonstrating its robust generation capabilities.}
\scalebox{0.75}{
\begin{tabular}{lccccccc}
\toprule
\textbf{Model} & \textbf{Cultural$\uparrow$} & \textbf{Time$\uparrow$} & \textbf{Space$\uparrow$} & \textbf{Biology$\uparrow$} & \textbf{Physics$\uparrow$} & \textbf{Chemistry$\uparrow$} & \textbf{Overall} \\
\midrule
\rowcolor{gray!15}
\multicolumn{8}{l}{Generating Only} \\
PixArt-$\alpha$~\cite{chen2023pixartalpha} & 0.45 & 0.50 & 0.48 & \textbf{0.49} & \textbf{0.56} & 0.34 & 0.47 \\
playground-v2.5~\cite{li2024playground} & \textbf{0.49} & \textbf{0.58} & 0.55 & 0.43 & 0.48 & 0.33 & 0.49 \\
SD-v1-5~\cite{Rombach_2022_CVPR} & 0.34 & 0.35 & 0.32 & 0.28 & 0.29 & 0.21 & 0.32 \\
SD-XL-base-0.9~\cite{podell2023sdxlimprovinglatentdiffusion} & 0.43 & 0.48 & 0.47 & 0.44 & 0.45 & 0.27 & 0.43 \\
FLUX.1-dev~\cite{labs2025flux1kontextflowmatching} & 0.48 & \textbf{0.58} & \textbf{0.62} & 0.42 & 0.51 & \textbf{0.35} & \textbf{0.50} \\
\midrule
\rowcolor{gray!15}
\multicolumn{8}{l}{Unified Understanding and Generation} \\
VILA-U~\cite{wu2025vilauunifiedfoundationmodel} & 0.26 & 0.33 & 0.37 & 0.35 & 0.39 & \textbf{0.23} & 0.31 \\
Janus-1B~\cite{chen2025janusprounifiedmultimodalunderstanding} & 0.16 & 0.26 & 0.35 & 0.28 & 0.30 & 0.14 & 0.23 \\
$\mathrm{ULM-R1^{\dagger}}$~\cite{jiang2025coreinforcementlearningunifiedmultimodal} & - & - & - & - & - & - & 0.33 \\
\rowcolor{cyan!5}
\textbf{\protect\methodname-1B} & \textbf{0.31} & \textbf{0.39} & \textbf{0.56} & \textbf{0.38} & \textbf{0.44} & 0.22 & \textbf{0.38} \\
\hdashline
Emu3~\cite{wang2024emu3nexttokenpredictionneed} & 0.34 & 0.45 & 0.48 & 0.41 & 0.45 & 0.27 & 0.39 \\
Janus-Pro-7B~\cite{chen2025janusprounifiedmultimodalunderstanding} & 0.30 & 0.37 & 0.49 & 0.36 & 0.42 & 0.26 & 0.35 \\
\rowcolor{cyan!5}
\textbf{\protect\methodname-7B} & \textbf{0.36} & \textbf{0.46} & \textbf{0.62} & \textbf{0.42} & \textbf{0.55} & \textbf{0.29} & \textbf{0.45} \\
\bottomrule
\label{tab:wise}
\end{tabular}}
\vspace{-1em}
\end{table*}

%% file: tables/geneval.tex
\begin{table*}[t]
\centering
\caption{\textbf{Results on GenEval.} PairUni achieves SOTA results among 1B models and competitive performance at 7B scale.
}
\scalebox{0.75}{
\begin{tabular}{lccccccc}
\toprule
\textbf{Method} & \textbf{Single Obj.} & \textbf{Two Obj.} & \textbf{Counting} & \textbf{Colors} & \textbf{Position} & \textbf{Color Attri.} & \textbf{Overall} \\
\midrule
\rowcolor{gray!15}
\multicolumn{8}{l}{Generating Only} \\
PixArt-$\alpha$~\cite{chen2023pixartalpha} & \textbf{0.98} & 0.50 & 0.44 & 0.80 & 0.08 & 0.07 & 0.48 \\
SDXL~\cite{podell2023sdxlimprovinglatentdiffusion} & \textbf{0.98} & 0.74 & 0.39 & \textbf{0.85} & 0.15 & 0.23 & 0.55 \\
DALL-E 3~\cite{poojadall} & 0.96 & \textbf{0.87} & \textbf{0.47} & 0.83 & \textbf{0.43} & \textbf{0.45} & \textbf{0.67} \\
\midrule
\rowcolor{gray!15}
\multicolumn{8}{l}{Unified Understanding and Generation} \\
SEED-X~\cite{ge2024seed} & 0.97 & 0.58 & 0.26 & 0.80 & 0.19 & 0.14 & 0.49 \\
Show-o~\cite{xie2024showo} & 0.95 & 0.52 & 0.49 & 0.82 & 0.11 & 0.28 & 0.53 \\
ILLUME~\cite{wang2024illume} & \textbf{0.99} & 0.86 & 0.45 & 0.71 & 0.39 & 0.28 & 0.61 \\
HermersFlow~\cite{yang2025hermesflowseamlesslyclosinggap} & 0.97 & 0.67 & \textbf{0.65} & 0.77 & 0.28 & 0.42 & 0.61 \\
UniRL~\cite{mao2025unirlselfimprovingunifiedmultimodal} & 0.95 & 0.74 & 0.27 & 0.81 & 0.62 & 0.52 & 0.65 \\
Janus-Pro-1B~\cite{chen2025janusprounifiedmultimodalunderstanding} & \textbf{0.99} & 0.82 & 0.48 & \textbf{0.90} & 0.62 & 0.57 & 0.73 \\
ULM-R1~\cite{jiang2025coreinforcementlearningunifiedmultimodal}  & - & - & - & - & - & - & 0.76 \\
Janus-Pro-R1~\cite{pan2025unlockingahamomentsreinforcement}  & 0.98 & 0.80 & 0.51 & 0.84 & 0.59 & 0.55 & 0.71 \\
\rowcolor{cyan!5}
\textbf{\protect\methodname-1B} & 0.98 & \textbf{0.91} & 0.44 & 0.75 & \textbf{0.95} & \textbf{0.69} & \textbf{0.79} \\
\hdashline
Chameleon~\cite{team2024chameleon} & -- & -- & -- & -- & -- & -- & 0.39 \\
D-DiT~\cite{li2025dual} & 0.97 & 0.80 & 0.54 & 0.76 & 0.32 & 0.50 & 0.65 \\
LWM~\cite{liu2025worldmodelmillionlengthvideo} & 0.93 & 0.41 & 0.46 & 0.79 & 0.09 & 0.15 & 0.47 \\
Transfusion~\cite{zhou2025transfusion} & -- & -- & -- & -- & -- & -- & 0.63 \\
TokenFlow-XL~\cite{qu2025tokenflow} & 0.95 & 0.60 & 0.41 & 0.81 & 0.16 & 0.24 & 0.55 \\
Janus-Pro-7B~\cite{chen2025janusprounifiedmultimodalunderstanding} & 0.97 & 0.88 & 0.57 & 0.90 & 0.77 & 0.64 & 0.79 \\
DSR~\cite{hong2025reinforcing} & -- & -- & -- & -- & -- & -- & 0.84 \\
Janus-Pro-R1~\cite{pan2025unlockingahamomentsreinforcement}  & \textbf{0.99} & \textbf{0.94} & 0.66 & 0.92 & 0.87 & \textbf{0.78} & \textbf{0.86} \\
\rowcolor{cyan!5}
\textbf{\protect\methodname-7B} & 0.97 & 0.75 & \textbf{0.78} & \textbf{0.97} & \textbf{0.91} & 0.69 & 0.85 \\
\bottomrule
\end{tabular}
}
\label{tab:gen_eval_results}
\vspace{-1em}
\end{table*}

%% file: figs/reward_curve.tex
\begin{wrapfigure}{r}{0.35\textwidth}
    \centering
    \includegraphics[width=0.8\linewidth]{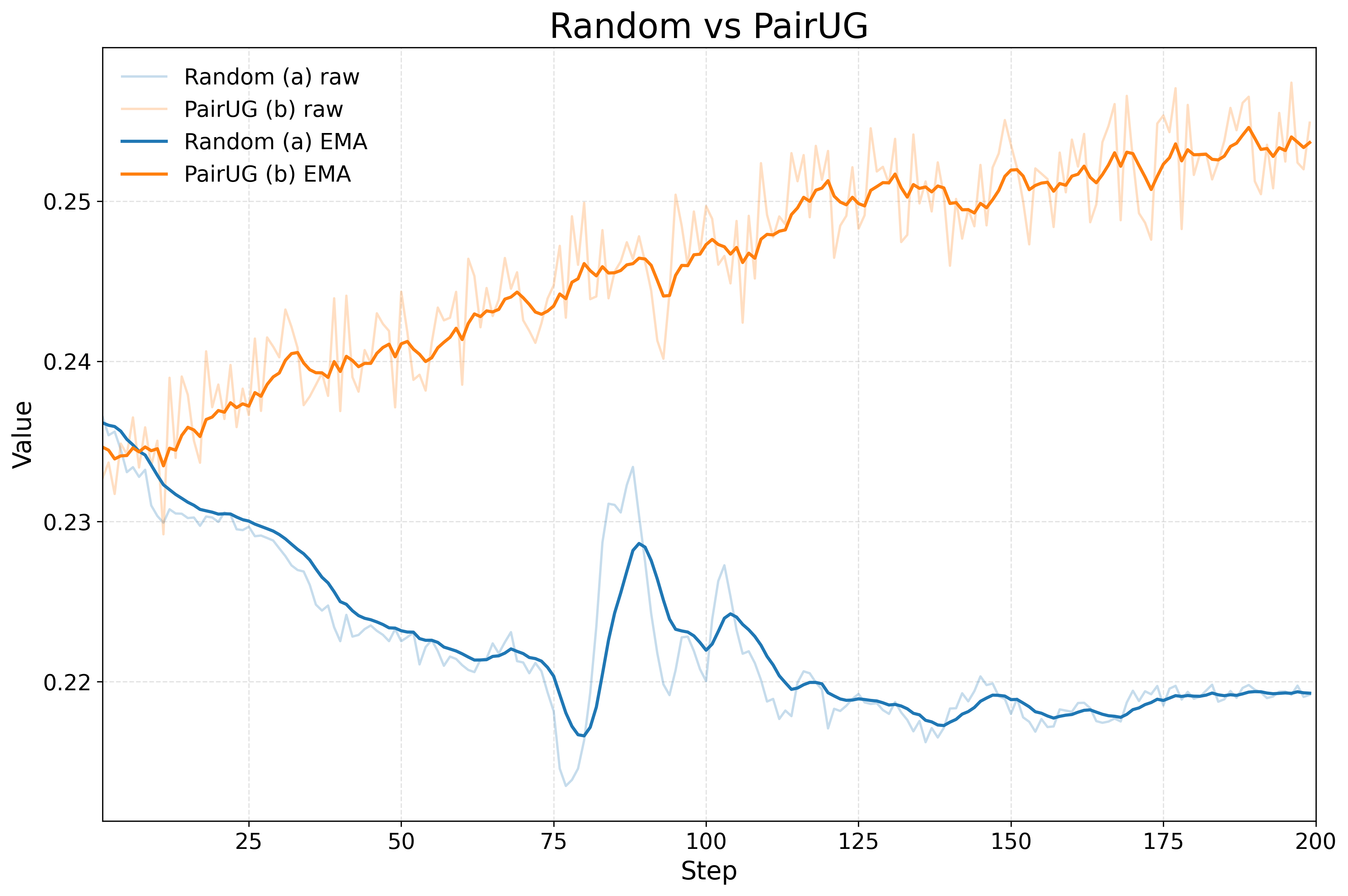}
    \caption{Comparison of training reward curves between PairUG-16k and random pairing strategies.}
    \label{fig:reward_curve}
\end{wrapfigure}

%% file: tables/more_model_performance.tex
\begin{table}[t]
\centering
\caption{\textbf{Extended Model Performance Analysis}: (a) Improvements with PairUni on Lumina-DiMOO. (b) Performance gains on Bagel with PairUG-16k. Add, Replace, Remove, and Style are subtasks of ImgEdit~\cite{ye2025imgedit}.}
\label{tab:more_model_performance}
\resizebox{0.85\columnwidth}{!}{
    \begin{tabular}{lccccccc}
        \multicolumn{8}{c}{\textbf{(a) Lumina-DiMOO Performance}} \\
        \toprule
        \textbf{Model} & \textbf{MMMU} & \textbf{MMStar} & \textbf{GenEval} & \textbf{Add} & \textbf{Replace} & \textbf{Remove} & \textbf{Style} \\
        \midrule
        Lumina-DiMOO~\cite{lumina-dimoo} & 58.6 & 52.4 & 0.88 & 3.82 & 3.83 & 2.76 & 4.18 \\
        w/ Random Pair & 56.1 & 52.1 & 0.86 & 3.75 & 3.66 & 2.75 & 3.98 \\
        \rowcolor{cyan!5}
        Lumina-DiMOO w PairUni & \textbf{61.3} & \textbf{52.6} & \textbf{0.89} & \textbf{3.84} & \textbf{3.94} & \textbf{2.78} & \textbf{4.22} \\
        \bottomrule
    \end{tabular}
}
\vspace{0.5em}

\resizebox{0.85\columnwidth}{!}{
    \begin{tabular}{lccccc}
        \multicolumn{6}{c}{\textbf{(b) Bagel Performance}} \\
        \toprule
        \textbf{Model} & \textbf{MMMU} & \textbf{POPE} & \textbf{GenEval (S)} & \textbf{GenEval (L)} & \textbf{ImgEdit (sum)} \\
        \midrule
        Bagel (report) & 55.3 & - & 82 & 88 & 3.38 \\
        Bagel (reproduced) & 52.8 & 87.37 & 77.9 & 86.1 & 3.38 \\
        w/ Random Pair & 51.3 & 87.62 & 81.7 & 86.3 & 3.39 \\
        \rowcolor{cyan!5}
        w/ PairUG-16k (Ours) & \textbf{54.0} & \textbf{89.21} & \textbf{84.6} & \textbf{87.2} & \textbf{3.51} \\
        \bottomrule
    \end{tabular}
}
\vspace{-1em}
\end{table}

%% file: tables/ablation_pair.tex
\begin{table}[t]
    \centering
    \begin{minipage}{0.48\textwidth}
        \centering
        \caption{\textbf{Ablation study of Data Pairing.} Comparison of different pairing strategies, highlighting the contribution of PairUG-16k.}
        \label{tab:data}
        \scalebox{0.65}{ 
            \begin{tabular}{lccc}
                \toprule
                \textbf{Model} & \textbf{MMMU} & \textbf{MMStar} & \textbf{GenEval} \\
                \midrule
                Pairs from $\mathcal{U}$ only & 38.2 & 43.7 & 0.75 \\
                Pairs from $\mathcal{G}$ only & 36.4 & 41.9 & 0.74 \\
                \hdashline
                Unpair  & 38.4 & 44.4 & 0.71 \\
                Random Pair & 38.4 & 44.3 & 0.73 \\
                Aligned-based Pairs & 39.2 & 44.6 & 0.76 \\
                Retrieval-based Pairs & 40.1 & 44.9 & 0.77 \\
                \hdashline
                PairUG (7.5K)  & 39.6 & 43.7 & 0.76 \\
                \rowcolor{cyan!5}
                PairUG-16k & \textbf{40.4} \improve{2.0} & \textbf{46.1} \improve{1.8} & \textbf{0.79} \improve{0.06} \\
                \bottomrule
            \end{tabular}
        }
    \end{minipage}
    \hfill 
    \begin{minipage}{0.48\textwidth}
        \centering
        \caption{\textbf{Ablation study on Trajectory-level credit assignment.}}
        \label{tab:sim}
        \scalebox{0.65}{
            \begin{tabular}{lcccc}
                \toprule
                \textbf{Model} & \textbf{MME(P)} & \textbf{MMMU} & \textbf{MMStar} & \textbf{GenEval} \\
                \midrule
                PairUni-1B w/o sim & 1469.87 & 40.0 & 45.1 & 0.79 \\
                \rowcolor{cyan!5}
                PairUni-1B & \textbf{1483.18} & \textbf{40.4} & \textbf{46.1} & 0.79 \\
                \hdashline
                PairUni-7B w/o sim & 1554.91 & 47.0 & 47.7 & 0.85 \\
                \rowcolor{cyan!5}
                PairUni-7B & \textbf{1597.71} & 47.0 & \textbf{49.5} & 0.85 \\
                \bottomrule
            \end{tabular}
        }
    \end{minipage}
\end{table}

%% file: sections/5-conclusion.tex
\section{Conclusion}
This paper introduces \methodname, a reinforcement learning framework for UVLMs  that aligns understanding and generation via paired training signals, and PairUG-16k, a curated dataset of understanding–generation pairs that supports consistent policy learning. On standard UVLMs evaluations with Janus‑Pro backbones, our approach achieves strong, balanced improvements in both understanding and generation, surpassing competitive RL baselines. 

%% file: sections/appendix_overview.tex
\noindent

\input{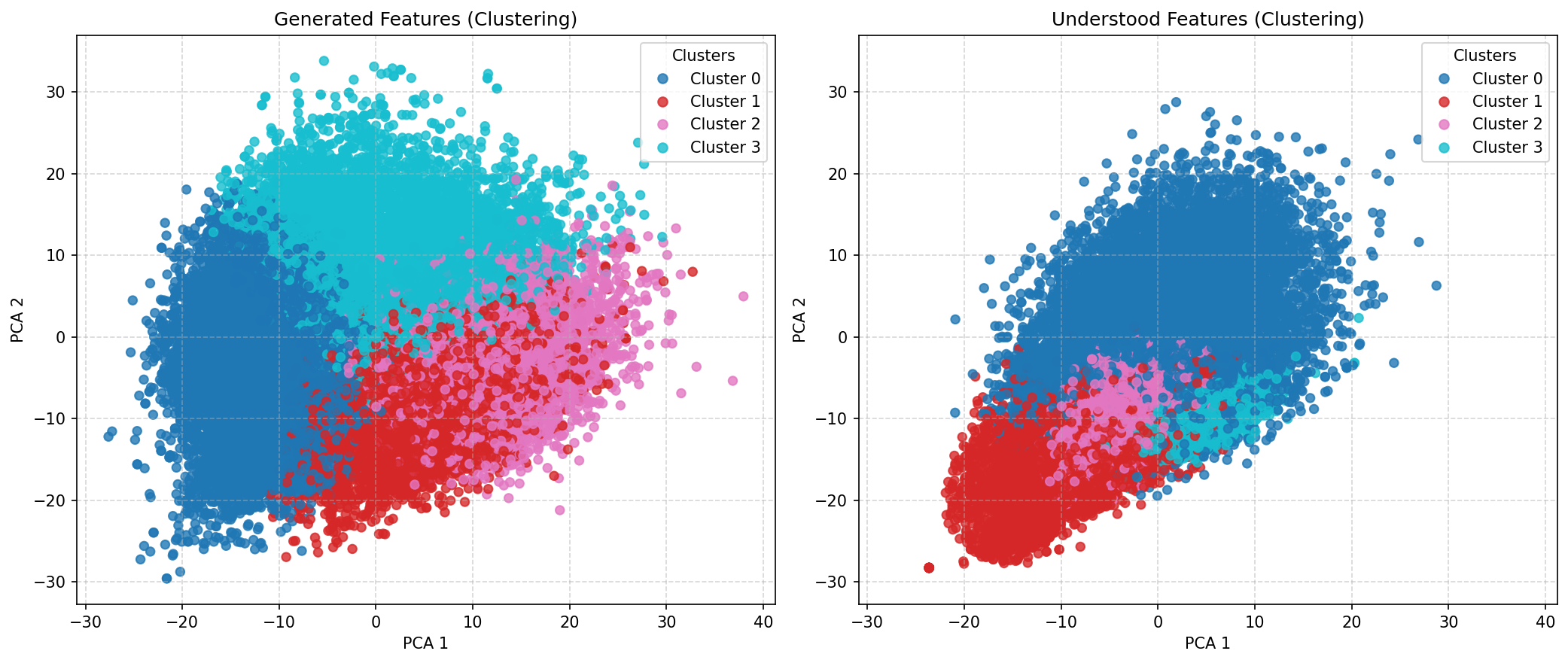}

\textbf{Overview.} 
Appendix~\ref{app:related_work} covers related work on UVLMs and the RL methods used in UVLMs.
Appendix~\ref{app:data} provides data-centric details, including summary statistics of the original data distribution, representative cases of retrieved pairs, and the empirical distribution of PairUG-16k.
Appendix~\ref{app:ablation} presents systematic ablation studies, including the choice of image extractor, weighting strategies, and threshold selection.
Appendix~\ref{app:more_model} demonstrates the generalization capabilities of our method on additional models.
Appendix~\ref{app:understanding-cases} presents qualitative case studies on understanding tasks.
Appendix~\ref{app:prompt} includes the prompts used by GPT-o3.
The closing sections include notes on the use of large language models.

%% file: figs/data_distribution.tex
\begin{figure}[ht]
    \centering
    \includegraphics[width=1.0\textwidth]{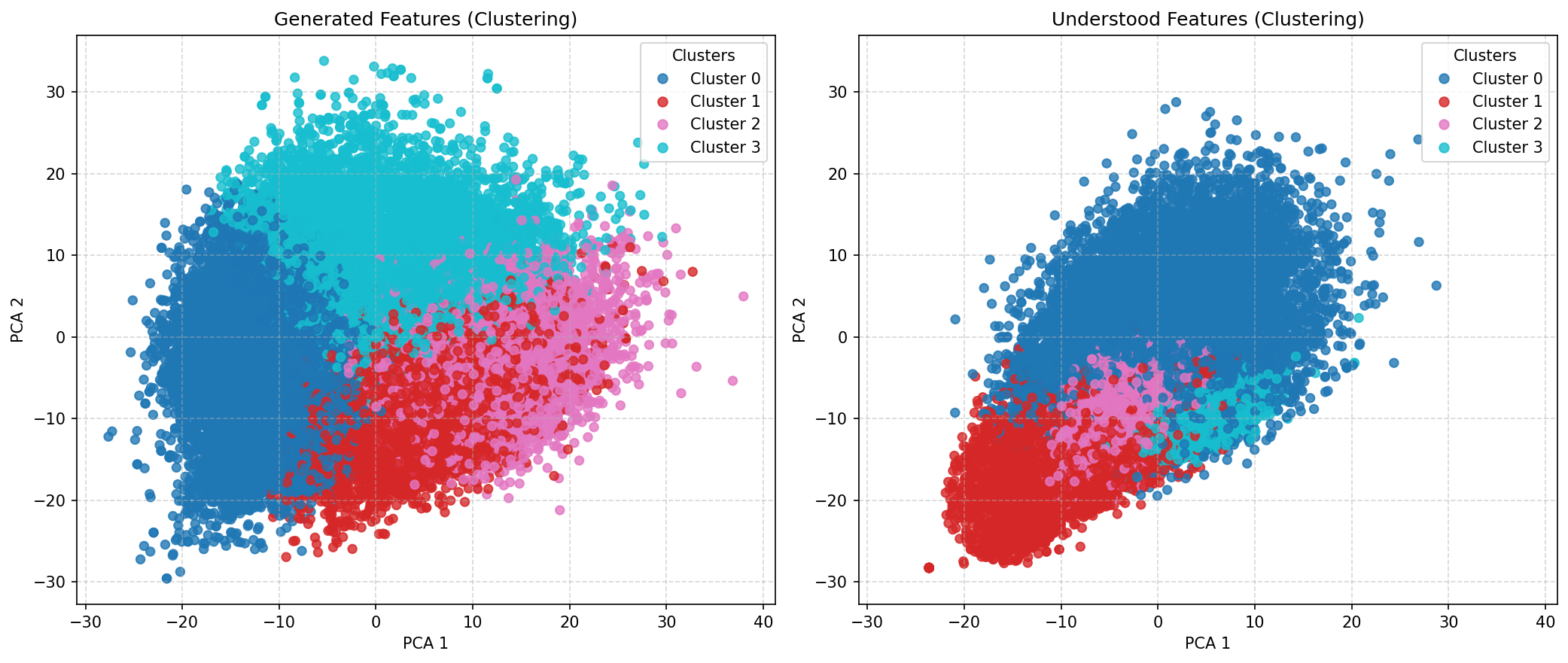}
    \caption{Distributional comparison between multimodal understanding and image generation datasets. The visualization highlights the domain gap and the necessity of the proposed alignment strategy.}
    \label{fig:data_distribution}
\end{figure}

%% file: sections/appendix_related_work.tex
\section{Related Work}
\label{app:related_work}

\noindent 
\textbf{Unified Vision-Language Model.}
The pursuit of unified multimodal frameworks has led to significant innovations in both architecture design and training paradigms. 
Early approaches~\cite{xie2024showo,team2024chameleon} like Show-o series~\cite{xie2024showo,xie2025showo2} establish the autoregressive foundations for joint vision-language processing.
Meanwhile, Transfusion~\cite{zhou2025transfusion} introduces diffusion-based methodologies to enhance generation quality. 
These foundational works, as systematically analyzed in~\cite{zhang2025unifiedmultimodalunderstandinggeneration}, demonstrate the potential of unifying modalities through shared representation learning.
Recent advances have pushed the boundaries of unified modeling~\cite{deng2025bagel,chen2025janusprounifiedmultimodalunderstanding}. 
For example, Janus-Pro~\cite{chen2025janusprounifiedmultimodalunderstanding} innovatively uses bidirectional encoder-decoder structures for understanding and generation, achieving stronger performance on both sides.
Bagel~\cite{deng2025bagel} adopts transformer experts and is trained with massive image generation and understanding data, leading to the state-of-the-art performance.
This architectural evolution aligns with the broader trend of developing modular yet integrated systems that can dynamically adapt to different modalities.
In this context, our work enhances UVLMs post-training, in particular, during the reinforcement learning phase. 
We present a novel view of pair data generation and utilization of such pair data with proposed PairGRPO.

\noindent
\textbf{Reinforcement Learning in UVLMs.} The integration of reinforcement learning (RL) has emerged as a critical component for advancing unified MLLMs during the post-training. 
Early RL applications focused on modality-specific enhancements: step-by-step rule-based rewards for mathematical reasoning~\cite{shao2024deepseekmathpushinglimitsmathematical}, and bbox IoU rewards for visual grounding~\cite{jiang2025t2i}. 
For text-to-image generation, CLIP-based rewards~\cite{radford2021learning} became standard for aligning visual outputs with textual descriptions. 
The paradigm shifted with unified RL approaches that exploit cross-modal synergies. 
T2I-R1~\cite{jiang2025t2i} pioneered iterative refinement through GRPO~\cite{shao2024deepseekmathpushinglimitsmathematical}, using detailed descriptions as intermediate rewards. 
Recently, several works also explore the RL-based post-training for UVLMs.
In particular, UniRL~\cite{mao2025unirlselfimprovingunifiedmultimodal} proposed a self-improving pipeline where generated QA pairs simultaneously serve as training data and reward signals, though this approach showed performance degradation in understanding benchmarks. 
More sophisticated reward mechanisms have since been developed. DSR~\cite{hong2025reinforcing} introduced dual-source rewards combining original image-caption pairs with generated content, while HermesFlow~\cite{yang2025hermesflowseamlesslyclosinggap} implemented Pair DPO~\cite{rafailov2024directpreferenceoptimizationlanguage} to enforce consistency between modalities. 
Notably, CoRL~\cite{jiang2025coreinforcementlearningunifiedmultimodal} adopts a two-stage approach, first training unified RL on shared data before specializing in understanding/generation phases, demonstrating improved performance across multiple benchmarks.
Different from these methods, which all focus on unified RL method design, our work provides a new view on understanding data and generation data.
We propose to build the UG pairs to benefit both tasks.
With proposed PairGRPO along with UG pairs, our work improves various UVLMs.

%% file: sections/appendix_data_details.tex
\section{More details about data}
\label{app:data}

\subsection{Data Distribution of Understanding and Generation Data}
\label{sec:data_distribution}
\input{figs/pair_data}
We curate two complementary splits covering multimodal understanding and image generation. For understanding, we adopt Orsta-47k~\cite{ma2025one}, a high-quality and diverse set that spans chart analysis, counting, object detection, grounding, mathematical reasoning, OCR, puzzles, and scientific reasoning. For image generation, BLIP-3o-60k~\cite{chen2025blip3} comprises curated AI-generated images paired with detailed textual descriptions, including single- and dual-object scenes as well as text-containing visuals. We deduplicate and ensure there is no overlap with the data used during pretraining.

To characterize their composition, we apply unsupervised clustering over the union of the two splits and examine cluster proportions (Figure~\ref{fig:data_distribution}). The two distributions exhibit pronounced divergence: categories prevalent in understanding data—such as math- or OCR-intensive items—are rare in the generation split, whereas descriptive object-centric scenes are comparatively overrepresented in generation.

\subsection{Retrieved Pairs Cases}
Figure~\ref{fig:pair_data} presents representative retrieved examples.

\subsection{Distribution of PairUG-16k}

\input{figs/pairug_distribution}

Figure~\ref{fig:PairUG-16k} summarizes the composition of PairUG-16k. We construct two complementary splits: \emph{Aligned Pairs} and \emph{Retrieved Pairs}.
For the Aligned Pairs, the data originate from two sources: generation (3{,}043 examples) and understanding (1{,}928 examples). The class distribution is long-tailed: the head classes---such as Math (646), Human Gestures (602), and Puzzle (402)---account for a substantial portion of the data, while several categories (e.g., object count with 65 instances) appear infrequently. This split provides high-quality supervision across 17 labeled categories.
The Retrieved Pairs display a different profile. The largest classes are JourneyDB (2{,}476) and GenEval (1{,}787), followed by Human Gestures (1{,}387), Object1 (1{,}232), Text2 (1{,}193), Text1 (992), Mscoco Human (857), and Occupation2 (708), with smaller categories such as Object2 (252) and Occupation1 (238). The similarity histogram is unimodal with most pairs in the 0.55--0.75 range, indicating that retrieval yields semantically related pairs while retaining diversity.

%% file: figs/pair_data.tex
\begin{figure}[t]
    \centering
    \includegraphics[width=0.85\textwidth]{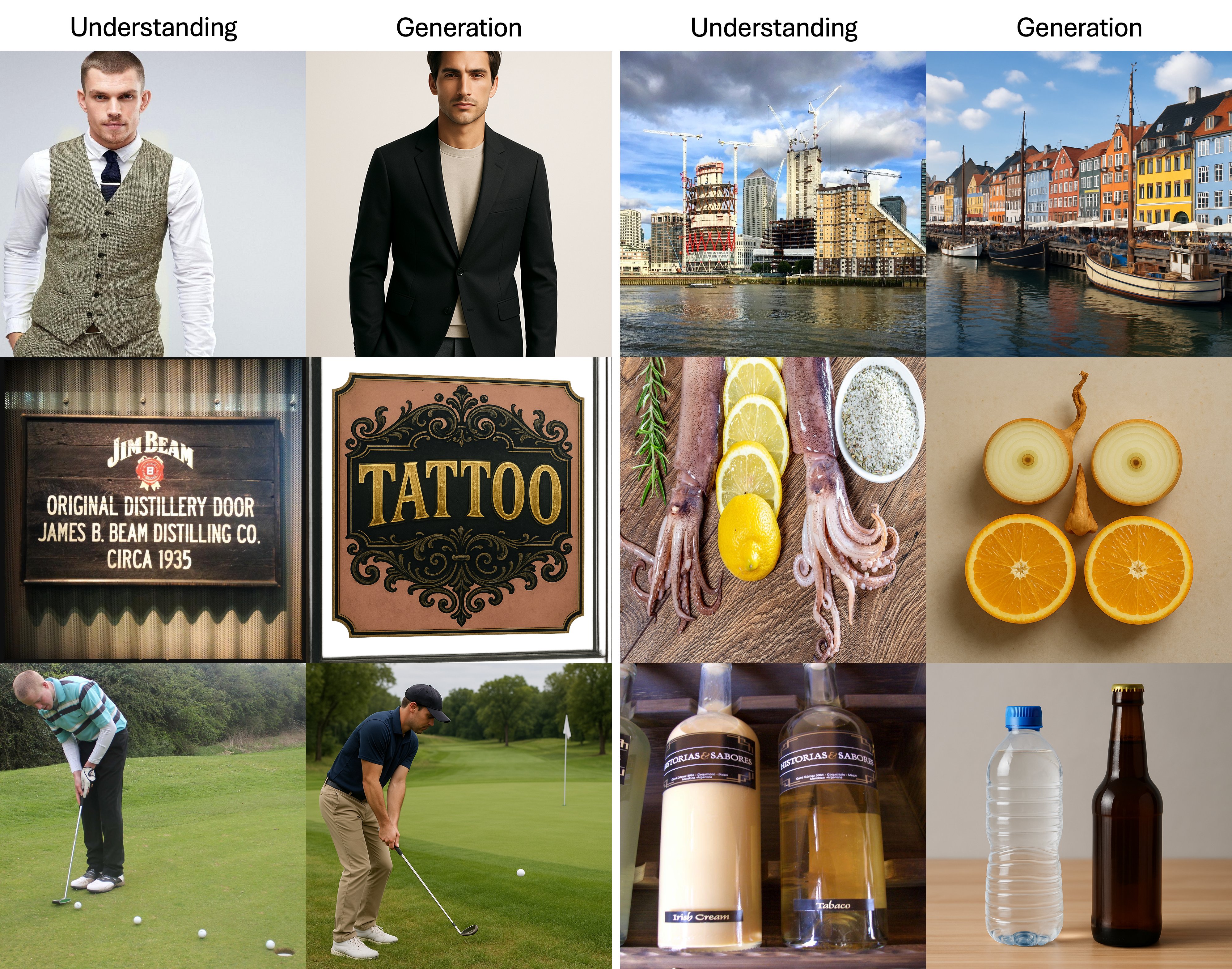}
    \caption{Representative paired cases demonstrating the synergy between multimodal understanding and image generation tasks. The examples highlight how aligned data pairs can enhance model performance across both domains.}
    \label{fig:pair_data}
\end{figure}

%% file: figs/pairug_distribution.tex
\begin{figure}[t]
    \centering
    \includegraphics[width=1.0\textwidth]{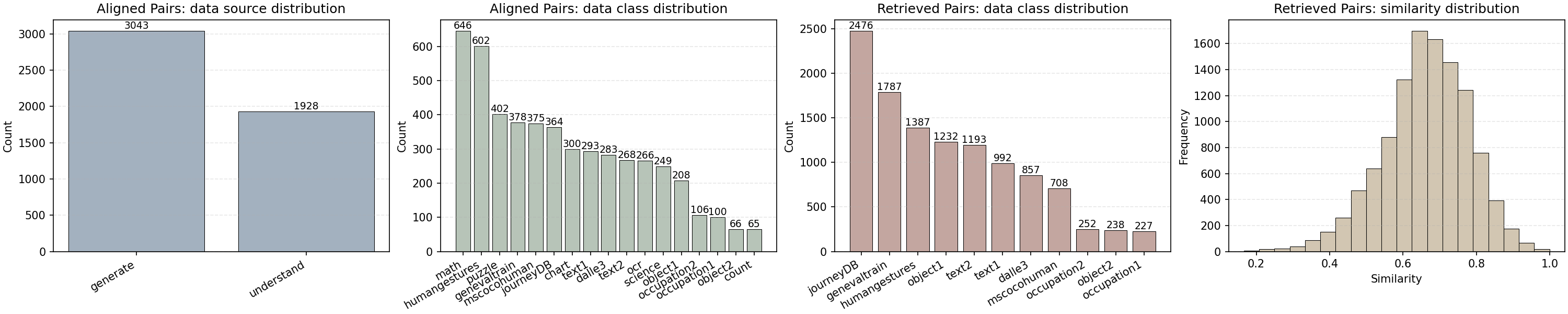}
    \caption{Detailed distribution analysis of the PairUG-16k dataset. From left to right: (1) Source breakdown showing the composition of Aligned Pairs; (2) Class distribution within Aligned Pairs; (3) Class distribution for Retrieved Pairs; and (4) Similarity score distribution for Retrieved Pairs, validating the diversity and quality of the constructed dataset.}
    \label{fig:PairUG-16k}
\end{figure}

%% file: sections/appendix_ablation.tex
\section{Additional Ablation Studies}
\label{app:ablation}

In this section, we provide detailed ablations on key design choices of \methodname, including the weighting strategy for retrieved pairs, the similarity threshold, the K-means medoid selection strategy, and the choice of image feature extractor. All experiments are conducted on the Janus-Pro-1B model.

\subsection{Weighting Strategy for Retrieved Pairs}
We compare different weighting schemes for the retrieved pairs in Eq. 2. As shown in Table~\ref{tab:ablation_weighting}, using the square root of the similarity score (`sqrt(similarity)`) provides the most robust performance boost. This is likely because it balances the contribution of high-similarity retrieval without dominating the loss.
\input{tables/ablation_weighting}

\subsection{Similarity Threshold Selection}
We investigate the impact of the similarity threshold used for filtering retrieved pairs. Table~\ref{tab:ablation_threshold} shows that a threshold of 0.6 is optimal. Lower thresholds (e.g., 0.5) introduce noise, which can degrade performance, while higher thresholds (e.g., 0.7) reduce the diversity of the training data and limit the benefits of retrieval augmentation.
\input{tables/ablation_threshold}

\subsection{K-means Medoid Selection}
We evaluate the effectiveness of using K-means medoids for data selection compared to random sampling. As demonstrated in Table~\ref{tab:ablation_kmeans}, using K-means medoids significantly outperforms random sampling. We hypothesize that outliers are not necessarily useless but often represent data points that cannot simultaneously adapt to both understanding and generation tasks (e.g., tabular data may be suitable for understanding but show a significant gap in generation performance). By using clustering and selecting medoids, we prioritize data points that are compatible with both tasks, yielding better performance than random sampling.
\input{tables/ablation_kmeans}

\subsection{Results with different image extractors.}
\label{app:image_encoder}
\input{tables/ablation_c}

We evaluate three alternatives for the image feature extractor: the Perception Encoder (PE)~\cite{bolya2025PerceptionEncoder}, DINOv3~\cite{simeoni2025dinov3}, and ResNet~\cite{koonce2021resnet}. The PE is designed for high-level semantic understanding, yet it underperforms on both the understanding and generation benchmarks (Table~\ref{tab:image_features}). In contrast, DINOv3 and ResNet---both emphasizing visual feature similarity---achieve comparable results. These findings indicate that, in our setting, enforcing consistency with respect to visual similarity is more critical than modeling semantic abstraction.

\section{Gradient Similarity Analysis}
\label{app:gradient_analysis}
Figure~\ref{fig:intro} illustrates the gradient conflict mechanism. We observe that when the data fed to understanding and generation are semantically aligned, the cosine similarity between their gradients increases. Higher gradient agreement correlates with stronger downstream results.
\input{figs/intro_analysis}

%% file: tables/ablation_weighting.tex
\begin{table}[t]
\centering
\caption{\textbf{Ablation study on the Weighting Strategy.} The table compares the base model (no weighting) with linear and square root similarity weighting schemes, demonstrating that the proposed square root weighting yields the best consistency and performance.}
\label{tab:ablation_weighting}
\scalebox{0.80}{
    \begin{tabular}{lcccc}
        \toprule
        \textbf{Weighting Strategy} & \textbf{MME (p)} & \textbf{MMMU} & \textbf{MMStar} & \textbf{GenEval} \\
        \midrule
        wo weighting (Base) & 1469.87 & 40.0 & 45.1 & 0.79 \\
        linear similarity & 1478.24 & 40.3 & 46.0 & 0.79 \\
        \rowcolor{cyan!5}
        \textbf{sqrt(similarity) (Ours)} & \textbf{1483.18} & \textbf{40.4} & \textbf{46.1} & \textbf{0.79} \\
        \bottomrule
    \end{tabular}
}
\end{table}

%% file: tables/ablation_threshold.tex
\begin{table}[t]
\centering
\caption{\textbf{Ablation study investigating the impact of the similarity threshold on model performance.} The table reports results for thresholds of 0.5, 0.6 (Ours), and 0.7 across MMMU, MMStar, and GenEval benchmarks.}
\label{tab:ablation_threshold}
\scalebox{0.80}{
    \begin{tabular}{lccc}
        \toprule
        \textbf{Similarity Threshold} & \textbf{MMMU} & \textbf{MMStar} & \textbf{GenEval} \\
        \midrule
        Sim $\ge$ 0.5 & 38.6 & 45.6 & 0.76 \\
        \rowcolor{cyan!5}
        \textbf{Sim $\ge$ 0.6 (Ours)} & \textbf{40.4} & \textbf{46.1} & \textbf{0.79} \\
        Sim $\ge$ 0.7 & 39.1 & 44.9 & 0.72 \\
        \bottomrule
    \end{tabular}
}
\end{table}

%% file: tables/ablation_kmeans.tex
\begin{table}[t]
\centering
\caption{\textbf{Ablation study on the K-means Medoid Selection strategy.} The results compare the performance of the proposed K-means based selection against a random selection baseline across multiple benchmarks.}
\label{tab:ablation_kmeans}
\scalebox{0.8}{
    \begin{tabular}{lccc}
        \toprule
        \textbf{Selection Strategy} & \textbf{MMMU} & \textbf{MMStar} & \textbf{GenEval} \\
        \midrule
        No K-means (Random) & 38.4 & 44.0 & 0.73 \\
        \rowcolor{cyan!5}
        \textbf{K-means (Ours)} & \textbf{40.4} & \textbf{46.1} & \textbf{0.79} \\
        \bottomrule
    \end{tabular}
}
\end{table}

%% file: tables/ablation_c.tex
\begin{table*}[t]
\centering
\caption{\textbf{Ablation study on Image Feature Extractors.} Comparison of model performance using different image feature extractors (PE, DINOv3, and ResNet). The results justify the choice of ResNet for the final implementation.}
\scalebox{0.80}{
\begin{tabular}{lcc|c}
\toprule
\textbf{Model} & \textbf{MMMU} & \textbf{MMStar} & \textbf{GenEval} \\
\midrule
PE~\cite{bolya2025PerceptionEncoder} & 40.1 & 45.5 & 0.77 \\
DINOv3~\cite{simeoni2025dinov3} & \textbf{40.4} & 46.0 & \textbf{0.79} \\
\rowcolor{cyan!5}
ResNet~\cite{koonce2021resnet} & \textbf{40.4} & \textbf{46.1} & \textbf{0.79} \\
\bottomrule
\end{tabular}}
\label{tab:image_features}
\end{table*}

%% file: figs/intro_analysis.tex
\begin{figure}[t]
    \centering
    \includegraphics[width=0.5\linewidth]{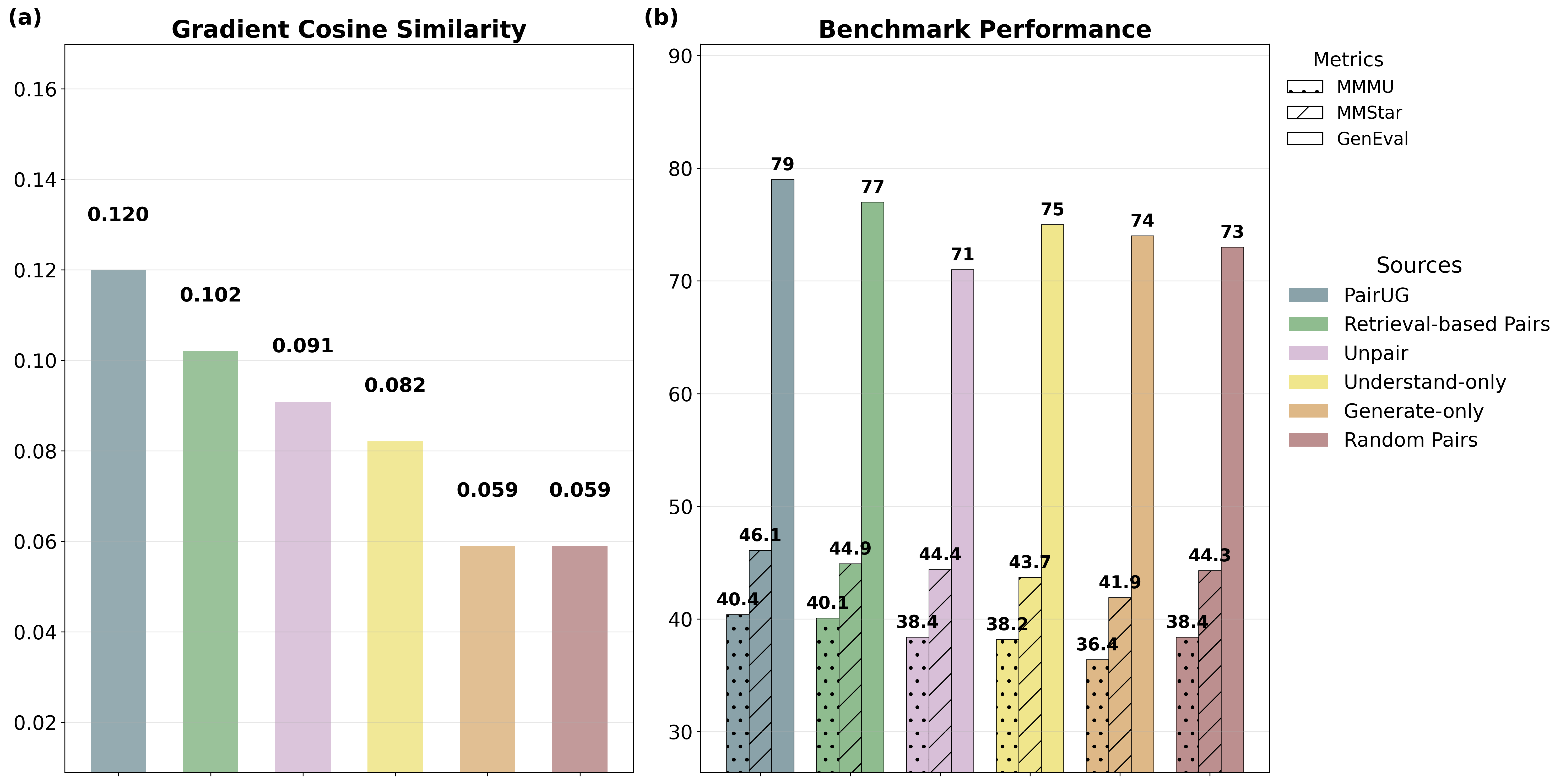}
    \caption{Performance Conflict Mechanism Analysis: Median gradient cosine similarity scores between understanding and generation components, alongside benchmark performance on two understanding benchmarks (MMMU~\cite{yue2024mmmu}, MMStar~\cite{chen2024we}) and one image generation benchmark (GenEval~\cite{ghosh2023geneval}). The analysis encompasses six distinct data combination scenarios: PairUG-16k, Retrieval-based Pairs, Unpair data with low similarity scores, pure Generation-only data, pure Understanding-only data, and Random Pairs.}
    \label{fig:intro}
\end{figure}

%% file: sections/appendix_more_models.tex
\section{Generalization Capabilities}
\label{app:more_model}

As shown in Table~\ref{tab:more_model_performance}, to assess the generalizability of our data and training method, we instantiate the PairGRPO framework on Lumina-DiMOO~\cite{lumina-dimoo}. Unlike Janus-Pro~\cite{chen2025janusprounifiedmultimodalunderstanding}, Lumina-DiMOO is a multimodal discrete diffusion model~\cite{yang2025mmada}. To enable parallel RL training for both understanding and generation, we set the rollout size to $2$ and adopt a fixed-step diffusion sampling scheme: we use $2$ diffusion steps for text generation and $35$ steps for image generation. We build upon the diffusion Large Language Model (dLLM)~\cite{nie2025largelanguagediffusionmodels} and its Proximal Policy Optimization (PPO)~\cite{wang2025trado} implementation, extending the framework to a multimodal dLLM and adding an implementation of the GPRO algorithm for policy optimization. Because Lumina-DiMOO has not released its paper nor the official training and evaluation code, this integration is preliminary and may benefit from further refinement. In the current implementation, PairUni improves both the model's comprehension and generation capabilities.

BAGEL \cite{deng2025bagel} is a strong unified vision–language model (UVLM) that integrates both understanding and generation components. It adopts a diffusion head~\cite{Adaptively-Diffusion,Siamese-Diffusion} with a flow-matching strategy for visual generation, together with an auto-regressive text-generation architecture. Since BAGEL does not support reinforcement learning (RL) training under its default settings, we perform supervised fine-tuning (SFT) using the PairUG-16k dataset. To satisfy the data format requirements of PairUG, the training data are constructed with a 1:1 ratio between understanding and generation tasks. For the generation task, the image resolution is set to 512, while all other hyperparameters follow the default configuration of BAGEL. The resulting SFT model achieves consistent performance improvements on the MMMU, POPE, GenEval (s) and GenEval (L), and ImgEdit benchmarks.

%% file: sections/appendix_case_studies_understanding.tex
\section{Case Studies on Understanding Tasks}
\label{app:understanding-cases}

\input{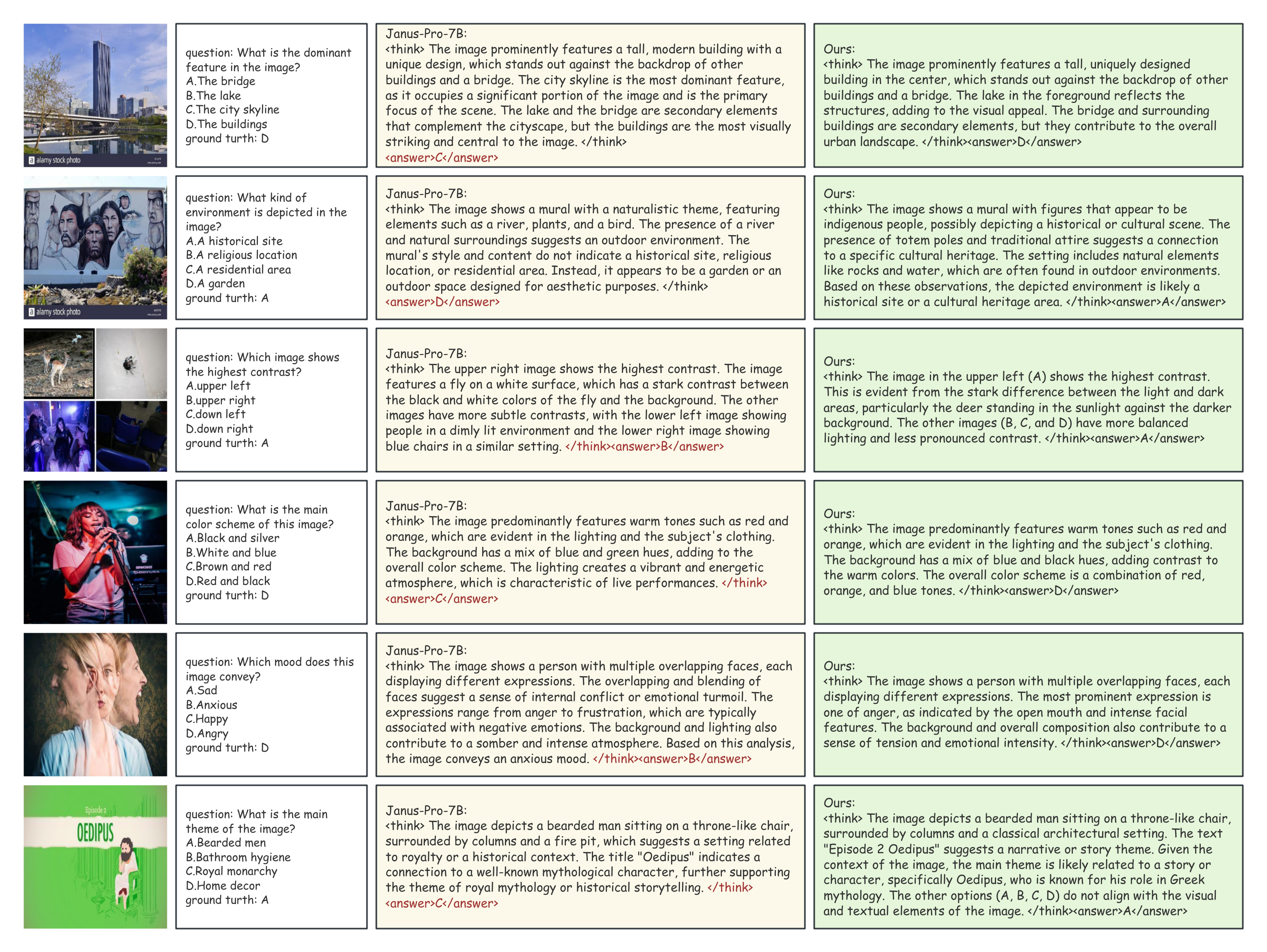}

As shown in Figure~\ref{fig:understand_case}, we present representative cases comparing Janus-Pro and PairUni on understanding tasks.

%% file: figs/understand_case.tex
\begin{figure}[htbp]
    \centering
    \includegraphics[width=\linewidth]{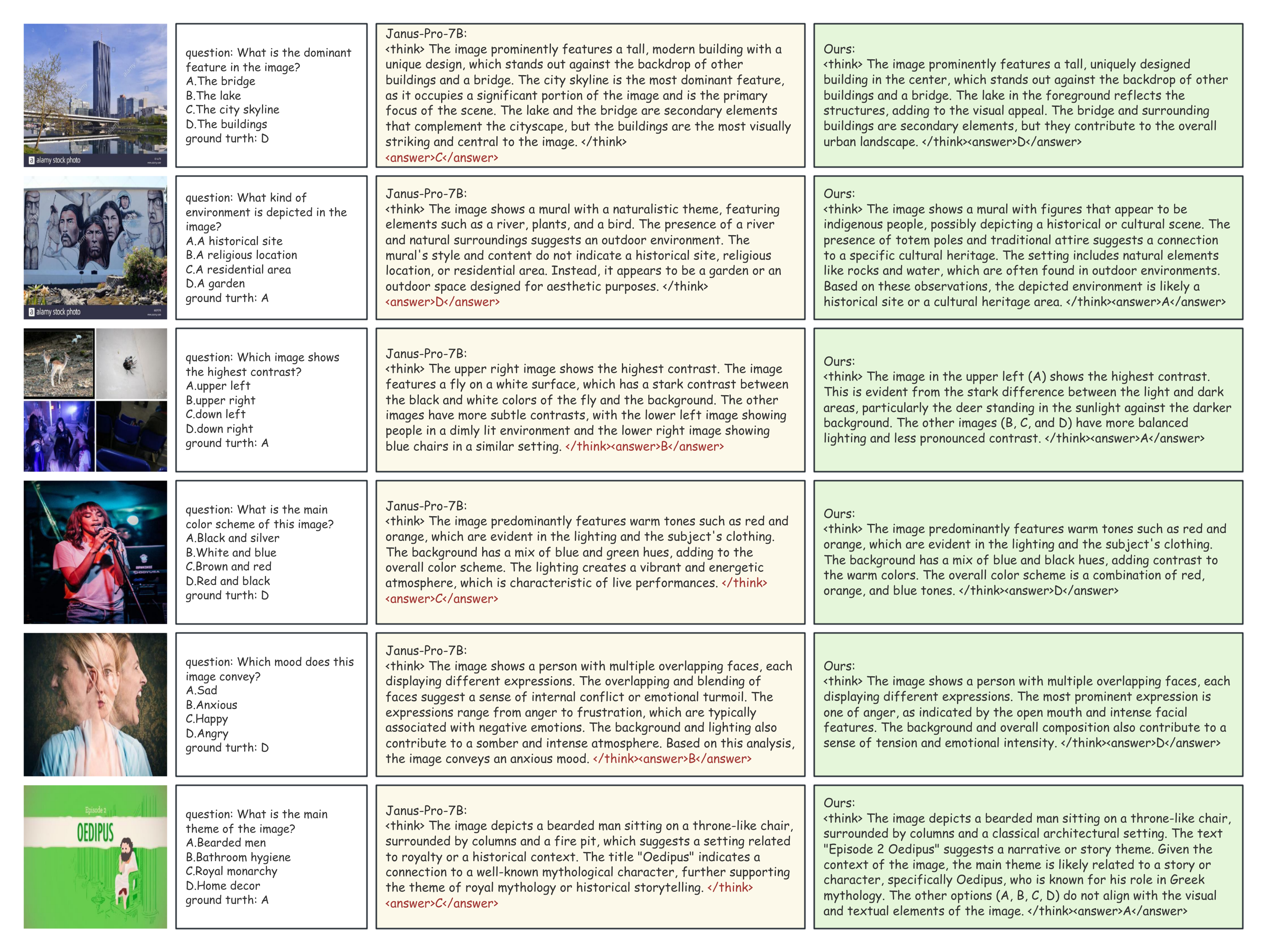}
    \caption{Representative qualitative examples comparing the multimodal understanding capabilities of Janus-Pro and PairUni. The cases highlight PairUni's improved accuracy in interpreting complex visual scenes.}
    \label{fig:understand_case}
\end{figure}

%% file: sections/appendix_prompts.tex
\section{Prompts for GPT-o3}
\label{app:prompt}
We design our instruction templates to enforce the \textit{semantic completion} logic described in Section~\ref{sec:data_pipeline}.
For the Understanding $\to$ Generation pathway, the prompt (Figure~\ref{fig:generate_prompt}) instructs GPT-o3 to generate a caption $C$ that is not only descriptive but also strictly consistent with the provided Question-Answer pair $(Q, A)$. It explicitly asks the model to verify that the visual evidence required for $Q$ is present in $C$.
For the Generation $\to$ Understanding pathway, the prompt (Figure~\ref{fig:understand_prompt}) directs the model to formulate questions $Q$ that target the key entities and attributes described in the caption $C$, ensuring the question is answerable given the generation intent.
Furthermore, both templates include a self-verification step where the model is asked to check for hallucinations or contradictions before finalizing the output. This rigorous prompting strategy ensures high-fidelity alignment within the constructed quadruples.

\input{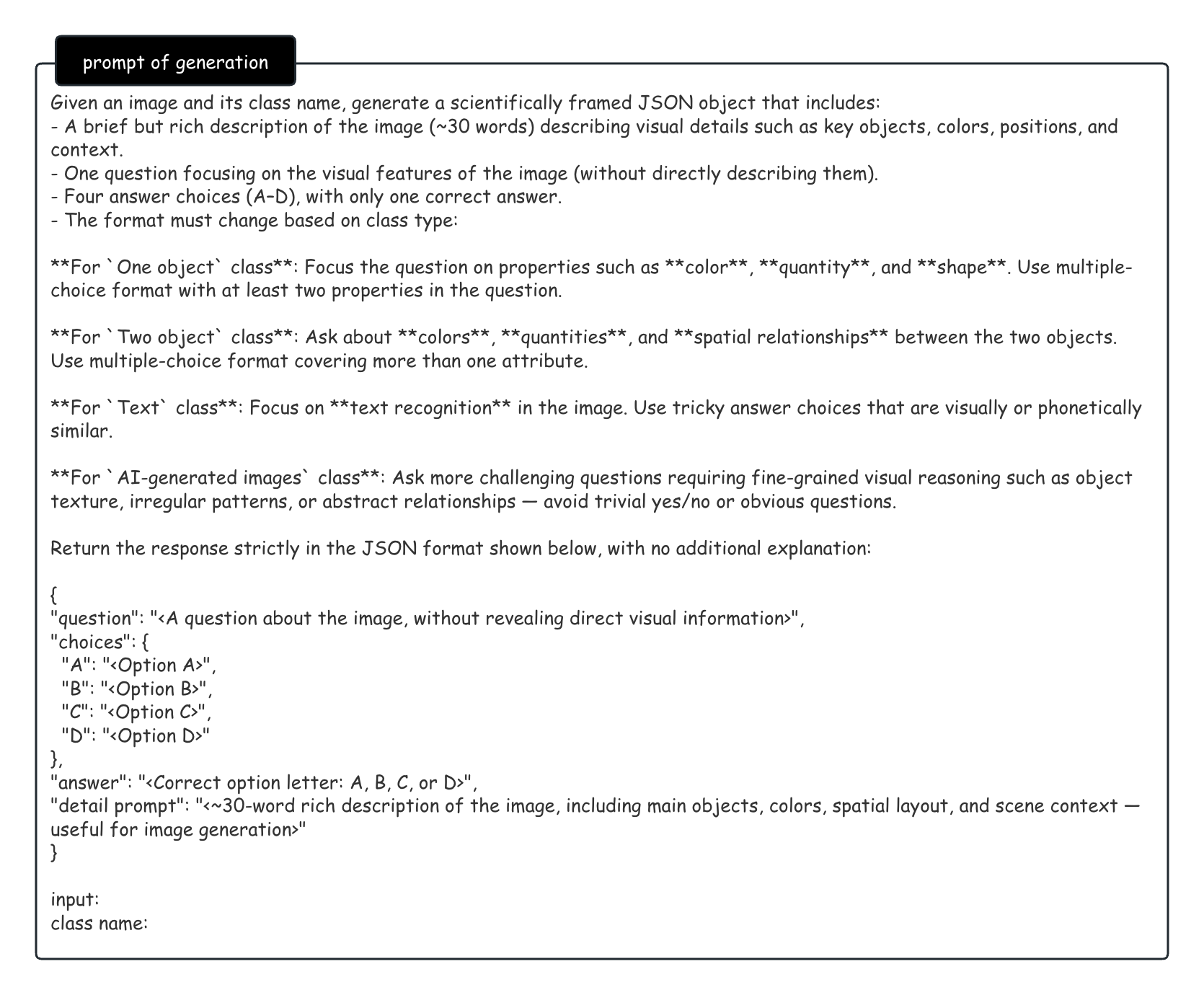}

\input{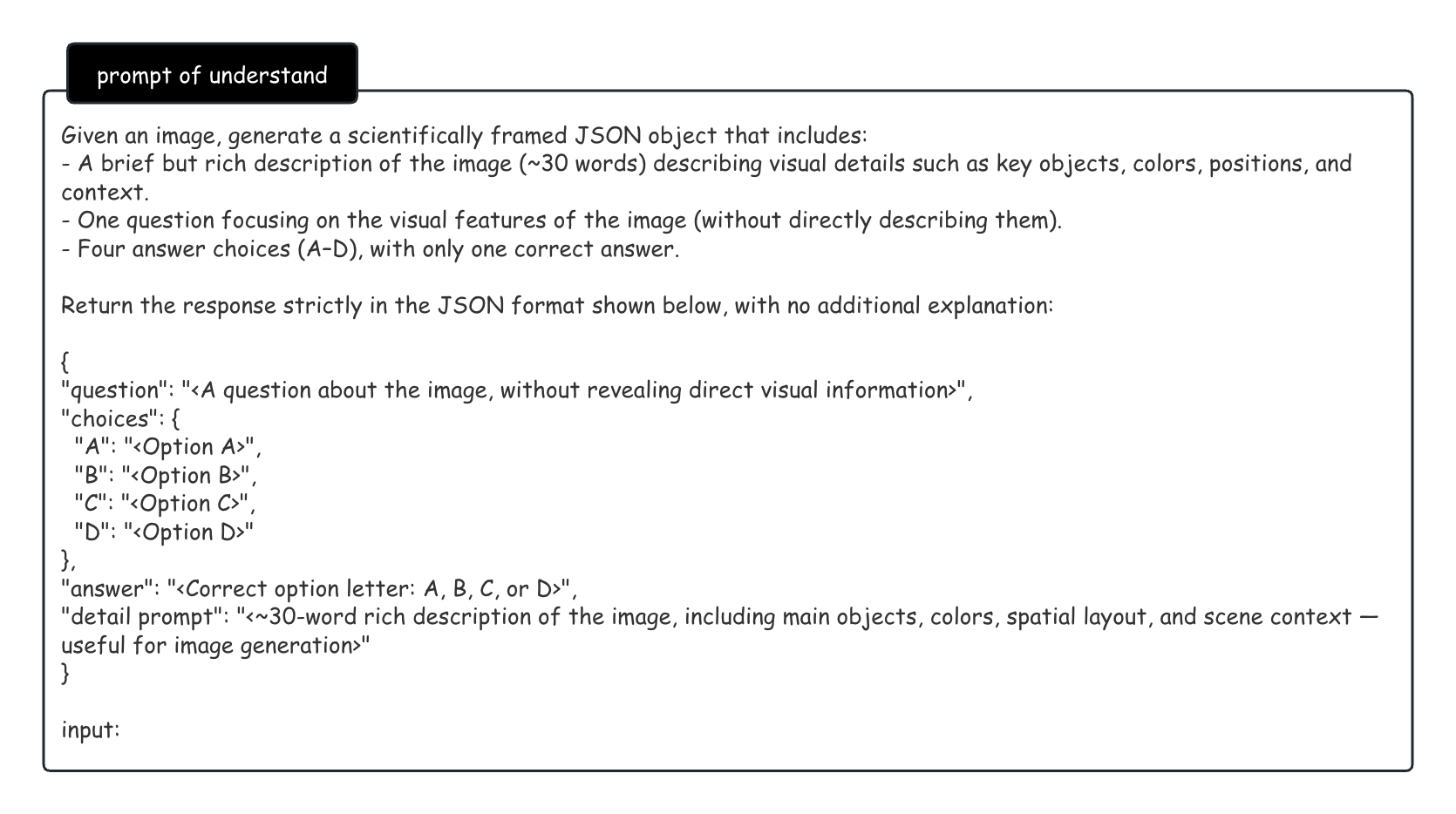}

%% file: figs/generate_prompt.tex
\begin{figure}[ht]
    \centering
    \includegraphics[width=1.0\textwidth]{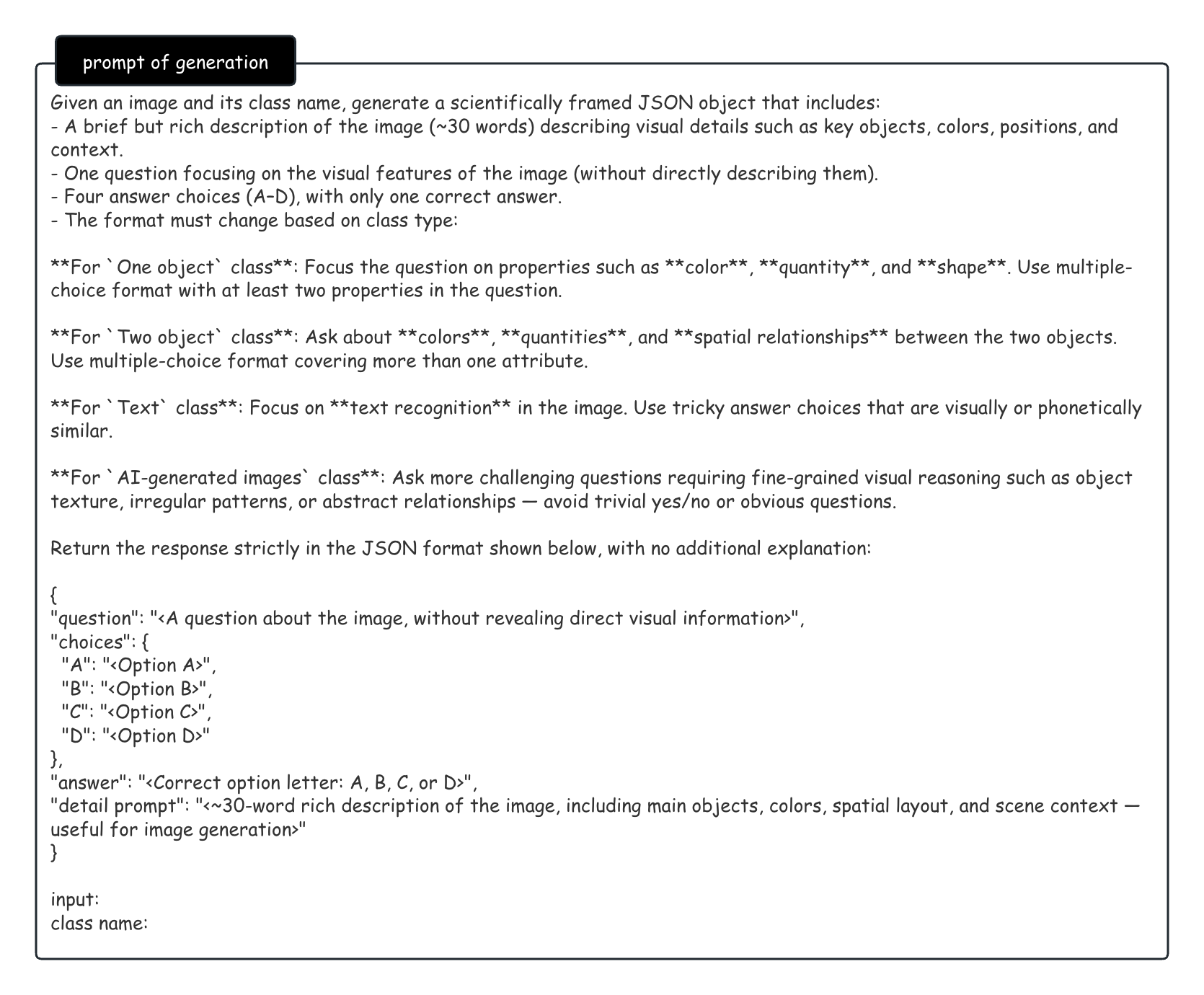}
    \caption{Detailed prompts employed for generating synthetic training data. These prompts are designed to ensure diversity and relevance in the generated image-text pairs.}
    \label{fig:generate_prompt}
\end{figure}

%% file: figs/understand_prompt.tex
\begin{figure}[ht]
    \centering
    \includegraphics[width=1.0\textwidth]{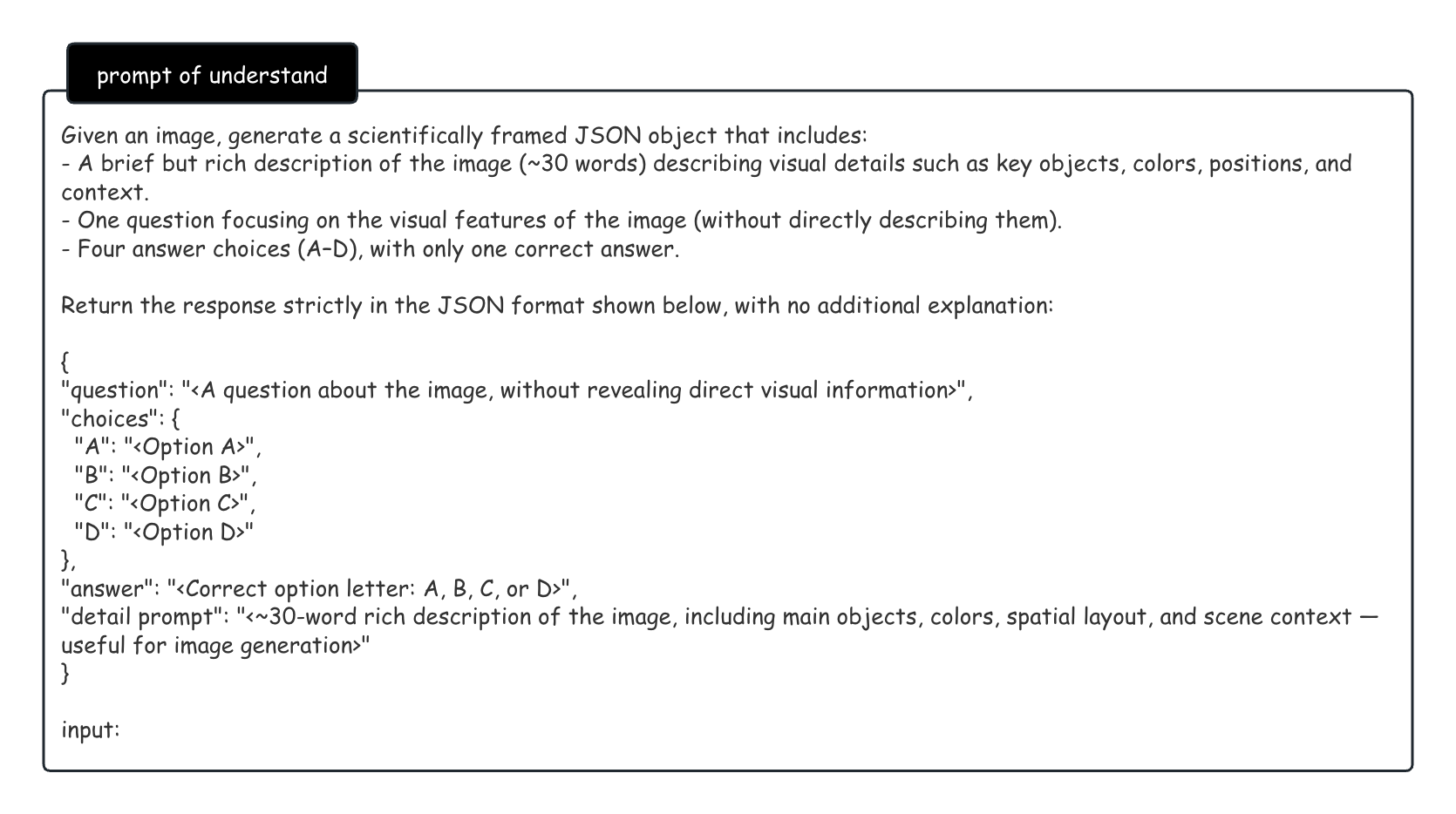}
    \caption{The prompt template used with GPT-o3 for constructing understanding-focused quadruple data. This prompt guides the model to generate relevant questions and answers based on the input image.}
    \label{fig:understand_prompt}
\end{figure}